# Physics-constrained robust learning of open-form partial differential equations from limited and noisy data

Mengge Du(都檬阁)[1], Yuntian Chen(陈云天)[2,4, a)], Longfeng Nie(聂隆锋)[3], Siyu Lou(楼思余)[4], Dongxiao Zhang(张东晓)[2,5,b)]

[1]*College of Engineering, Peking University, Beijing, 100871, P.R. China*

[2]*Ningbo Institute of Digital Twin, Eastern Institute of Technology, Ningbo, 315200, Zhejiang, P.R. China*

[3]*School of Environmental Science and Engineering, Southern University of Science and Technology, Shenzhen 518055, P. R. China*

[4]*Eastern Institute for Advanced Study, Eastern Institute of Technology, Ningbo, 315200, Zhejiang, P.R. China*

[5]*National Center for Applied Mathematics Shenzhen (NCAMS), Southern University of Science and Technology, Shenzhen, 518000, Guangdong, P. R. China*

Corresponding author
Email address: a) ychen@eitech.edu.cn, b) zhangdx@eitech.edu.cn.

## ABSTRACT

Unveiling the underlying governing equations of nonlinear dynamic systems remains a significant challenge. Insufficient prior knowledge hinders the determination of an accurate candidate library, while noisy observations lead to imprecise evaluations, which in turn result in redundant function terms or erroneous equations. This study proposes a framework to robustly uncover open-form partial differential equations (PDEs) from limited and noisy data. The framework operates through two alternating update processes: discovering and embedding. The discovering phase employs symbolic representation and a novel reinforcement learning (RL)-guided hybrid PDE generator to efficiently produce diverse open-form PDEs with tree structures. A neural network-based predictive model fits the system response and serves as the reward evaluator for the generated PDEs. PDEs with higher rewards are utilized to iteratively optimize the generator via the RL strategy and the best-performing PDE is selected by a parameter-free stability metric. The embedding phase integrates the initially identified PDE from the discovering process as a physical constraint into the predictive model for robust training. The traversal of PDE trees automates the construction of the computational graph and the embedding process without human intervention. Numerical experiments demonstrate our framework's capability to uncover governing equations from nonlinear dynamic systems with limited and highly noisy data and outperform other physics-informed neural network-based discovery methods. This work opens new potential for exploring real-world systems with limited understanding.

## I. INTRODUCTION

For an extended period, the discovery of scientific knowledge has been largely obtained on the first principle. With the advancement of experimental science and the exponential increase in data, data-driven methods have attracted increasing attention. There is an urgent desire to efficiently extract physical knowledge from data, particularly governing equations, to enhance our understanding of the natural world.

Many linear and nonlinear dynamic systems are described by parsimonious PDEs. Consequently, this has led to the emergence of an important branch in the realm of knowledge discovery, namely, PDE discovery. The primary task of PDE discovery is to identify governing equations directly from data. The governing equations considered in most relevant studies can be represented as:

$$u_t = F(\Theta(u,x);\xi) \tag{1}$$

where $u$ is the state variable of interest. The left-hand side (LHS) of the equation is a known function term represented by the time derivative of $u$. The target is to find an explicit nonlinear function $F$ on the right-hand side (RHS) of the equation, composed of a library of function terms and their corresponding coefficients. Derived from the Sparse Identification of Nonlinear Dynamic Systems (SINDy) approach [1], PDE functional identification of nonlinear dynamics (PDE-FIND) has brought about a significant breakthrough in the field of PDE discovery [2]. It proposes an algorithm named sequential threshold ridge regression to discover the linear combinations of function terms from measurements on a regular grid. Specifically, PDE-FIND signifies a static and predefined candidate library, consisting of monomial terms formed by $u$ and its partial derivatives, which is shown below:

$$\Theta(u,x) = [1, u, \partial_x u, u(\partial_x u), u^2, (\partial_x^2 u), ..., u^m(\partial_x^n u)] \tag{2}$$

where $m$ and $n$ represent the highest polynomial degree of $u$ and the highest order of its partial derivatives, respectively. Eq (1) can then be transformed into

$$u_t = \Theta \xi = (1, u, \partial_x u, ..., u^m(\partial_x^n u)) \bullet (\xi_1, \xi_2, \xi_3, ..., \xi_{m \times n}) \tag{3}$$

where the majority of coefficients $\xi$ are relegated to zero to preserve the sparsity of the identified equation. Although PDE-FIND successfully uncovers many canonical dynamics systems, such as the Navier-Stokes equation and the reaction-diffusion equation, there are two significant limitations: (1) The construction of the library hinges heavily on strong assumptions and prior knowledge, which is not available and applicable to unknown systems. A sufficiently large or overcomplete library is often required, which is impractical in many cases; (2) The method relies on numerical differentiation to assess partial derivatives on a regular grid, a practice that proves vulnerable to noise and sparse data, particularly in the context of high-order derivatives. Consequently, relevant methods can only discover equations within known domains with high-quality data; however, the identification of equations from real-world nonlinear systems remains a challenging task.

Without sufficient prior knowledge, it is difficult to construct a complete library that can cover all components of the PDE to be discovered. Meanwhile, the computational cost of a large library is also unacceptable for sparse regression. Subsequent studies have proposed methods based on an expandable library and even discovered open-form equations. Essentially, the expandable library is capable of producing new equation terms through the interaction of predefined basis function terms. Based on the symbolic neural network proposed in Equation

Learner (EQL) [3], PDE-Net 2.0 generates a flexible combination of operations and state variables by adjusting the network topology[4]. Compared with SINDy, it boasts a more compact library, and the computational cost is significantly reduced. Deep learning and genetic algorithm (DLGA) [5] and evolutionary partial differential equations (EPDE) [6] integrated the genetic algorithm (GA) to substantially expand the original candidate set through the recombination of gene fragments (via mutation and crossover operations). Nevertheless, the expressive power of the interactive space remains limited. Only multiplication and addition operations are introduced in the generation of interactive function terms, which is insufficient to discover open-form equations. Actually, in the early studies, many researchers tried to uncover analytical relations from nonlinear dynamic systems through symbolic mathematics [7,8]. Symbolic genetic algorithm (SGA) further combined symbolic representations and GA and represented PDEs by a collection of binary trees [9]. This method greatly increases representation flexibility, but the introduction of crossover and mutation operations may lead to poor iterative stability of the generated equations. To counter this, the following research further utilized the neural-guided agent to generate PDEs and accelerate the optimization process with deep reinforcement learning [10]. Note that symbolic mathematics enables the discovery of open-form PDEs, thus circumventing the high-memory consumption and computational requirements of overcomplete library methods.

The evaluation of state variables and their corresponding derivatives is vulnerable to noisy and sparse observations. The imprecise evaluation can result in nonnegligible coefficient errors or even incorporating redundant function terms in the identification process of PDEs. Some research has concentrated on tackling the second problem from various perspectives, such as employing the weak-formulation method [11–13] or Bayesian estimation [14–17] to deal with noisy and limited data. A significant branch of this research field is based on the neural networks-based method. Specifically, these algorithms tended to utilize deep neural networks to approximate the noisy system response, thereby allowing for the seamless evaluation of derivatives through automatic differentiation [18–20]. With the development of physics-informed neural network (PINN) [21,22], some methods tried to explicitly incorporate the loss term of physics into the neural network training process to enhance the robustness to noisy observations. A framework named DeepMod marks a significant breakthrough, which put forward a fully differentiable method for revealing the underlying governing equations with the training paradigm of PINN [23]. In this framework, a new loss term, composed of the residuals between the linear combinations of library terms and the time derivative (i.e., LHS) is integrated into the training process. The coefficients of function terms are kept sparse by the constraint of $L_1$ regularization and the final function terms are ascertained via a hard threshold. Similar to this method, PINN-SR employed sparse regression to evaluate the coefficients and executed STRidge along with the training of the neural networks using an alternating direction optimization strategy (ADO) strategy [24]. In subsequent studies, methods have been further refined; for instance, PDE_READ substituted the regular fully connected network with two rational neural networks [25], and Thanasutives et al. incorporated the discrete Fourier transform (DFT) in a multi-task learning paradigm for data denoising [26]. Although methods combined with PINN augment the ability to extract correct equations from noisy data, it is apparent that these methods rely on a predefined compact library based on prior knowledge. This is due to the computational inefficiency of embedding all of the function terms into the physics loss. A sufficiently large library poses more challenges in identifying the parsimonious equation. Fine-tuning the hyperparameters of sparsity for different systems and noise levels proves unreasonable and impractical. Xu et al. [27] employed GA to identify the initial equations and manually incorporated the discovered equations into the program to complete the physical constraint embedding process. However, this method leads to the discontinuity of PDE discovery and PINN-type training, thereby increasing the demand for human intervention.

Considering the aforementioned challenges, we propose a framework named R-DISCOVER, a **R**obust version based on prior work named **D**eep **I**dentification of **S**ymbolically **C**oncise **O**pen-form PDEs **V**ia **E**nhanced **R**einforcement-learning (DISCOVER) [10], which is aimed at simultaneously solving the two problems present in PDE-FIND. Firstly, we utilize symbolic representations to uncover open-form PDEs. A PDE can be represented by a binary tree instead of a pre-defined complete basis function library. Moreover, a neural network (NN) is used to fit and smooth observations, and the equations discovered are embedded as new physical constraints to handle noisy and sparse data. Particularly, the process of identifying the underlying equations via the proposed R-DISCOVER comprises two primary procedures: the discovery process and the embedding process. The discovery process first utilizes a RL-guided hybrid PDE generator to batch-generate diverse PDE expressions. Then, a NN-based predictive model is built to fit observations and provide predictions at any given location. It also serves as a reward evaluator to evaluate the performance of generated PDE expressions based on automatic differentiation. Finally, the initially identified PDE is selected from the generated better-fitting PDEs by a model selection method by balancing the data fitness and coefficients stability. During the embedding process, the initially identified PDEs are automatically incorporated as a physics loss into the NN by traversing the PDE tree. Subsequently, with the discovered physical information incorporated, the predictions derived from the NN and the evaluations of PDE samples can effectively minimize the interference caused by noise, thus becoming more precise. The alternating training of these two processes completes the closed loop of knowledge discovery and embedding, with the primary objective of uncovering underlying equations from limited and noisy data.

The main contributions of the proposed framework are mainly as follows:

- It seamlessly integrates symbolic mathematics and automatic equation embedding to robustly discover open-form PDEs from nonlinear systems. With the discovered equation as a physical constraint, it can effectively deal with highly noisy and sparse data.
- It is capable of generating diverse PDE expressions with a tree structure and efficiently completing the optimization process. A RL-guided hybrid PDE generator is designed by combining the RL, GA, and short-term memory of frequent function terms named dynamic subtree bank (DSB). The generator can produce complex PDE expressions even including multiple state and spatial variables in lengthy form. Mathematical and physical constraints, such as spatial symmetry, are considered to reduce the search space.
- We propose a parameter-free model selection method based on bootstrapped data, which strikes a balance between data fitness and stability of coefficients. The proposed metric is parameter-free and exhibits robustness across different scenarios.

The numerical experiments demonstrate that our framework can correctly uncover governing equations from linear and nonlinear dynamic systems even when faced with highly noisy and limited data. Its robustness and accuracy surpass other related PINN-based discovery methods.

The structure of this paper is organized as follows: We elucidate the specific operations of the framework, focusing on equation discovery in Sec. II and automatic PDE constraint embedding in Sec. II B. In Sec. III, we demonstrate the framework's capability to reveal governing equations under conditions of limited and noisy data through six numerical experiments and provide a comparative analysis with related methods. Lastly, in the conclusion, we present our primary findings and discuss potential avenues for future research.

## II. METHODOLOGY

The framework aims to leverage symbolic representation to find the analytical physical relationship between symbols. Assuming the target system of interest includes state variable $u$, and spatial-temporal variables $x$ and $t$, we believe the form of $\Theta(u,x)$ is as follows:

$$\Theta(u,x) = [f_1(u,x), f_2(u,x), ..., f_n(u,x), ...]. \tag{4}$$

This kind of representation liberates the constraints of a fixed candidate library. It can not only deal with high-order derivatives and high-degree polynomials, but also enables the discovery of more complex equation structures, such as fractional ($u/x$) and compound structures ($\partial_x^2(u^2)$). Assuming that the accessible noisy and sparse measurements can be denoted as $U_m = \{\tilde{u}(t_j, x_j)\}_{j=1}^{N_m} \subseteq (0,T] \times \Omega$, and sampled collocation points are represented as $U_c = \{\hat{u}(t_i, x_i)\}_{i=1}^{N_c} \subseteq (0,T] \times \Omega$, our goal is to directly extract the explicit form of the governing equations from them.

### A. Overview of the proposed framework

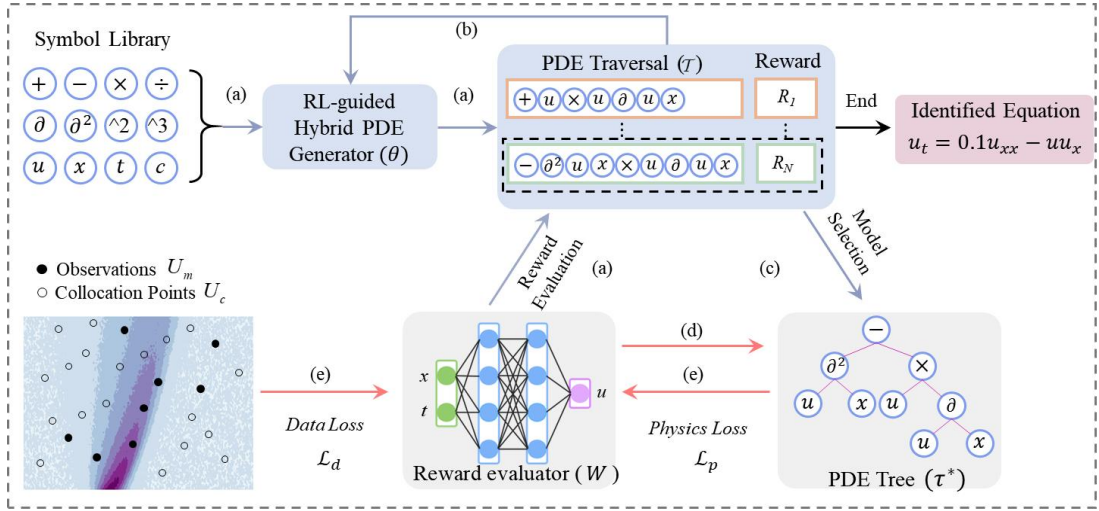

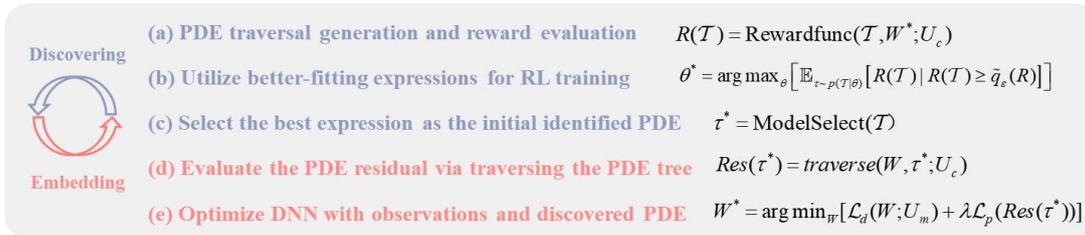

(a) PDE traversal generation and reward evaluation   $R(\mathcal{T}) = \text{Rewardfunc}(\mathcal{T}, W^*; U_c)$

**Discovering**

(b) Utilize better-fitting expressions for RL training   $\theta^* = \arg\max_\theta \left[ \mathbb{E}_{\tau \sim p(\mathcal{T}|\theta)}[R(\mathcal{T}) \mid R(\mathcal{T}) \geq \tilde{q}_\varepsilon(R)] \right]$

(c) Select the best expression as the initial identified PDE   $\tau^* = \text{ModelSelect}(\mathcal{T})$

(d) Evaluate the PDE residual via traversing the PDE tree   $Res(\tau^*) = traverse(W, \tau^*; U_c)$

**Embedding**

(e) Optimize DNN with observations and discovered PDE   $W^* = \arg\min_W [\mathcal{L}_d(W; U_m) + \lambda \mathcal{L}_p(Res(\tau^*))]$

**FIG. 1.** Overview of R-DISCOVER for discovering Burgers' equation from limited and noisy data. The whole framework consists of 5 procedures: (a) PDE traversal generation and reward evaluation; (b) Utilize better-fitting expressions for RL training; (c) Select the best expression as the initial identified PDE; (d) Evaluate the PDE residual via traversing the PDE tree; (e) Optimzie DNN with observations and discovered PDE.

To effectively mine correct governing equations from limited and noisy data, we propose the R-DISCOVER framework as illustrated in FIG. 1. The workflow of our proposed framework consists of two primary procedures: discovering and embedding. During the discovering process (Sec. II B) which corresponds to (a), (b), and (c) as shown in FIG. 1, the objectives are to explore and generate as many high-reward PDE representations as possible and determine the initial PDE that best matches the current data and physics information. We first preset a symbol library and utilize an RL-guided hybrid PDE generator to generate PDE expressions (FIG. 1 (a)). These expressions are represented in the form of PDE traversal, specifically, pre-order traversal of the corresponding PDE tree. The hybrid PDE generator utilizes an enhanced Long Short-Term Memory (LSTM) agent proposed in DISCOVER [10] to batch-generate initial samples. Then those samples are further processed by GA and short-term memory of high-quality function terms named DSB to produce more diverse samples. Subsequently, we build a NN-based predictive model to fit system responses. The predictive model can evaluate the system response and partial derivatives at any location within the computational domain. Therefore, it also serves as a reward evaluator to calculate the reward of generated PDE expressions. The LSTM agent is then updated by the risk-seeking policy gradient method with selected better-fitting expressions to provide a better initial PDE population[28] (FIG. 1 (b)). Finally, we perform model selection based on the refined PDE candidates, thoroughly evaluating data fitness and the stability of coefficients (FIG. 1 (c)). The optimal PDE is chosen as the initially identified PDE for subsequent physics embedding in the NN-based reward evaluator.

The second step corresponds to the embedding process (Sec. II C), as depicted in FIG. 1 (d) and (e). During the (d) process, the initially identified PDE can be expressed as the tree structure. By using the automatic differentiation of the reward evaluator, we can calculate the equation's residual while traversing the PDE tree and then obtain the physical loss. In this computation process, the physical constraint is automatically embedded into the reward evaluator without manually constructing the computational graph with code. Combined with supervised learning of observations, the reward evaluator can be trained and optimized similarly to PINN, which corresponds to the (e) process. The main purpose of step 2 is to utilize the preliminarily identified equations as physical constraints so that they can enhance the robustness to noise and evaluate the rewards of PDE samples in the discovering stage more accurately. Overall, these two steps are complementary processes, with the former mining underlying physical information from the data, and the latter applying PDE constraints to better correct the data information in the mining process. The strategy of alternating iterative updates like ADO[24] is implemented to obtain the final accurate and concise governing equation.

To sum up, a reinforcement learning-based framework is proposed to discover governing equations without prior information. It utilizes a hybrid generator to efficiently generate diverse open-form PDEs and automatically embeds the discovered PDE into neural networks to handle noisy and sparse data. A parameter-free metric is designed to determine the correct PDEs.

## B. PDE Discovering process

In this section, we will provide a detailed description of the discovery process, including the representation, generation, evaluation, optimization, and selection of the PDE.

### 1. Symbolic representations for open-form PDEs

To uncover the open-form governing equation underlying the nonlinear dynamic systems, we adopt a symbolic representation to express the equation. First, we need to define a symbol library consisting of two categories of symbols, i.e., operators ( $\{+,-,\times,\div,\partial,\partial^2,\wedge 2,...,etc.\}$ ) and operands ( $\{u,x,t,...,etc.\}$ ). For example, if we consider the right-hand side of Burgers' equation, which is $u_{xx}+uu_x$, it can be represented in several related forms as shown in Table I. A PDE can be represented by a binary tree structure, where non-leaf nodes are operators and leaf nodes are operands. A binary operator needs two operators to satisfy the requirement of full degrees, such as '+'. For a unary operator, it requires only one operand. Based on these syntactic relationships, a PDE tree and its pre-order traversal are in one-to-one correspondence. Furthermore, a PDE tree can be further partitioned into smaller-level representations, i.e., function terms or subtrees, according to the '+' and '-' operators on the top of trees. It's worth noting that to simplify the process, we do not consider the coefficients of the equation terms during the generation process. These coefficients can be evaluated during the reward evaluation phase.

**TABLE I.** Illustration of different forms of PDE and its compositions.

| Mathematical expression | PDE expressions | | Function terms | Subtree |
|---|---|---|---|---|
| | PDE Traversal | PDE tree | | |
| $u_{xx}+uu_x$ | $\{+\partial^2 ux \times u\partial ux\}$ | (tree diagram) | $\{u_{xx}, uu_x\}$, $\{\partial^2 ux, \times u\partial ux\}$ | (subtree diagram) |

### 2. Hybrid PDE traversal generator

Next, we provide a more detailed explanation of the PDE generation process. Given the complexity of generating a PDE expression tree, the previous work DISCOVER, employed a neural-based sequence generator to efficiently batch-generate corresponding PDE pre-order traversals [10]. Specifically, an enhanced structure-aware LSTM serves as the agent. By leveraging structured input in conjunction with monotonic attention, the agent effectively captures structured information, making it suitable for handling PDE traversal sequences with inherent structural attributes. Note that while the neural-based agent can generate PDE expressions in parallel and can be updated efficiently with the RL strategy, there are two significant challenges that need to be dealt with: (1) If the generated samples at the beginning stage keep at a low-level reward, those bad samples would in turn make the optimization fall into local optima. It's tricky to balance the exploration and exploitation in the training process if only the RL-strategy is utilized to optimize the generated equations; (2) Compared to an overcomplete library, free combinations of symbols result in a larger search space, which significantly increases the training burden, especially for those lengthy PDEs in multi-variables and high dimensional systems with limited understanding.

To address these issues, our framework proposes a RL-guided hybrid PDE generator by further integrating genetic operations from GA and a short-term memory of frequent function terms (DSB). Firstly, GA employs crossover and mutation operations, which have been proven to possess strong global search capability[29–31]. These operations can ensure the diversity of the generated samples, and avoid the problem that the samples generated by the agent tend to be consistent with the optimization process. The integration of GA and RL has been demonstrated to effectively solve global optimization and improve search efficiency in various optimization problems[29,32,33]. Secondly, at each iteration, the framework collects high-quality samples from GA and the RL agent, and conserves frequent function terms as prior information to form a dynamic subtree bank. Samples can be generated directly from the linear combinations of components in DSB, which significantly reduces the difficulty of optimization. The incorporation of DSB enables the proposed framework to address the challenges of equation discovery in complex systems with limited understanding The hybrid generation can leverage the advantages of each method, reducing the number of iterations while being more conducive to the discovery of complex and lengthy governing equations. As shown in FIG. 2 (a), the generated PDE traversals consist of three parts:

- **RL set**: The PDE samples from the reinforcement-learning set are batch-generated by the neural-based agent parameterized by $\theta$ in a step-by-step manner. Particularly, at $i^{th}$ iteration, the token $\tau_i$ is sampled from the symbol library on the probability distribution $p(\tau_i | \tau_{1:(i-1)}; \theta)$ emitted by the LSTM agent. The likelihood of a complete pre-order traversal can then be represented by $p(\tau | \theta) = \prod_{i=1}^{|\tau|} p(\tau_i | \tau_{1:(i-1)}; \theta)$. Physical and mathematical constraints are also embedded by modifying the library distribution to reduce the search space and ensure the generation of reasonable expressions. For example, the left child node following a partial differential operator cannot be a spatial variable (such as *x*). More details can be found in [10].

- **GA set:** The initial population of the genetic algorithm set is derived from high-quality samples in the RL set with higher rewards. As illustrated in ① of FIG. 2, through the crossover and mutation operations in GA, new offspring are produced and constitute the GA set. Crossover operates on function terms, while mutation targets tree nodes, i.e., operands or operators. This process mainly mimics the transmission and expression of genes in a population, increasing the diversity of equation representations. It's worth noting that in each iteration of GA, the initial population undergoes a restart operation based on the samples generated by RL, which is a more favorable approach compared to random initialization.

- **Expressions from DSB:** First, high-quality samples generated from both RL and GA are selected and each sample is split into subtrees (function terms). Then, frequent terms are selected from the pool of subtrees to form a basis function library, which is called DSB. At each iteration, the DSB will be updated and perform as a short-term memory of high-quality function terms. Common sparse regression is then utilized to generate new PDE expressions with the optimization objective $\hat{\xi} = \arg\min_\xi \|\Theta\xi - u_t\|_2^2 + \lambda\|\xi\|_0$, as shown in ② of FIG. 2.

Note that the $l_0$ penalty (λ) is not a fixed value. Instead, it is dynamically adjusted based on the mean squared error (MSE) of the RHS and LHS of the PDE expression identified up to the current iteration with the highest reward. λ can be expressed by $\lambda = \partial MSE(\Theta^*\xi^*, u_t)$, where $i$ refers to the current iteration and $\partial$ is randomly sampled from a preset interval, such as $\partial \in [0.1, 0.5]$, allowing for the generation of PDE expressions with different levels of sparsity penalty. Meanwhile, the function term library $\Theta$ is also subsampled from the DSB to further increase the diversity of generation.

---

**Algorithm 1** Generation of PDE expressions by RL-guided hybrid PDE generator
---
1: **Input**: LSTM agent with parameter $\theta$; Size of initial population $N$; Size of dynamic subtree bank $N_f$; Size of subsampled basis function library $N_{sub}$; Basis function library split from the best expression $\Theta^*$.
2: $\mathrm{T}_{RL} \leftarrow \{\tau_i \sim p(\cdot|\theta)\}_N^{i=1}$  // Initiate PDE traversals with a LSTM agent.
3: $\mathrm{T}_m \leftarrow Mutation(\mathrm{T}_{RL})$  // Generate N traversals through mutation.
4: $\mathrm{T}_c \leftarrow Crossover(\mathrm{T}_{RL})$  // Generate 2N traversals through crossover.
5: $\mathrm{T}_{GA} = \mathrm{T}_m \cup \mathrm{T}_c$  // Merge the expressions generated by GA.
6: $\mathrm{T} = \mathrm{T}_{RL} \cup \mathrm{T}_{GA} = \{\tau_i\}_N^{i=1}$  // Filter and keep best $N$ expressions.
7: Select frequent function terms from $\mathrm{T}$ to form a dynamic subtree bank.
$$\Psi = \{\psi_i\}_{i=1}^{N_f}$$
8: **For** $i$ in $1, 2, ... N$ **do**
9:   $\partial = Random(0.1, 0.5)$  // Randomly generate a coefficient for the sparse penalty.
10:   $\Theta_i = subsample(\Psi) = \{\psi_i\}_{i=1}^{N_{sub}}$  // Subsample DSB and form basis function library.
11:   $\lambda = \partial MSE(\Theta^*\hat{\xi}^*, u_t)$  // Generate a $l_0$ penalty.
12:   $\hat{\xi} = \arg\min_\xi \|\Theta_i\xi - u_t\|_2^2 + \lambda\|\xi\|_0$
13:   Select function terms with large coefficients to form PDE traversal and add it to the candidate set:
$$\mathrm{T}_{DSB} \leftarrow \tau_i \leftarrow \{\psi_j : |\hat{\xi}_j| > tol, j \in 1, 2, ..., N_{sub}\}$$
14: **End**
15: **Output**: PDE expressions for training $\mathrm{T}_{train} = \mathrm{T} \cup \mathrm{T}_{DSB}$

The RL-guided hybrid PDE generator essentially leverages a learnable agent to generate a good initial population, expands the combinations among the high-quality expressions through GA, and then confluences the high-quality genes through a dynamic subtree bank. The whole generation process is outlined in Algorithm 1 and can be found in the source code.

Spatial symmetry in domain knowledge is also effectively incorporated into the generation process. Specifically, during the sample generation process, sparse regression is employed to lessen the complexity of searching for equation representations in discrete space. Note that for multidimensional nonlinear systems, there often exist multiple spatial derivatives, such as $\{\omega_x, \omega_y, \omega_{xx}, \omega_{yy}\}$ in Navier-Stokes (NS) equations, which typically exhibit symmetrical relationships in spatial dimensions, i.e., the number of occurrences of $x$ and $y$ should be consistent. However, it is challenging for the PDE expressions to obey this constraint in the generation stage. To solve this, our algorithm further introduces a spatial correction to ensure the symmetry properties of physics. For example, the PDE expressions generated by the above method may appear as irregular cases such as $u\omega_x + v\omega_y + \omega_{xx}$. By introducing two correction methods of addition and deletion, we can transform it into $u\omega_x + v\omega_y$ and $u\omega_x + v\omega_y + \omega_{xx} + \omega_{yy}$, making it more physically meaningful and avoiding the generation of invalid forms. The hybrid generator, complemented by spatial symmetry constraints, enables our framework to have the ability to deal with complex governing equations from nonlinear dynamic systems even with multiple variables.

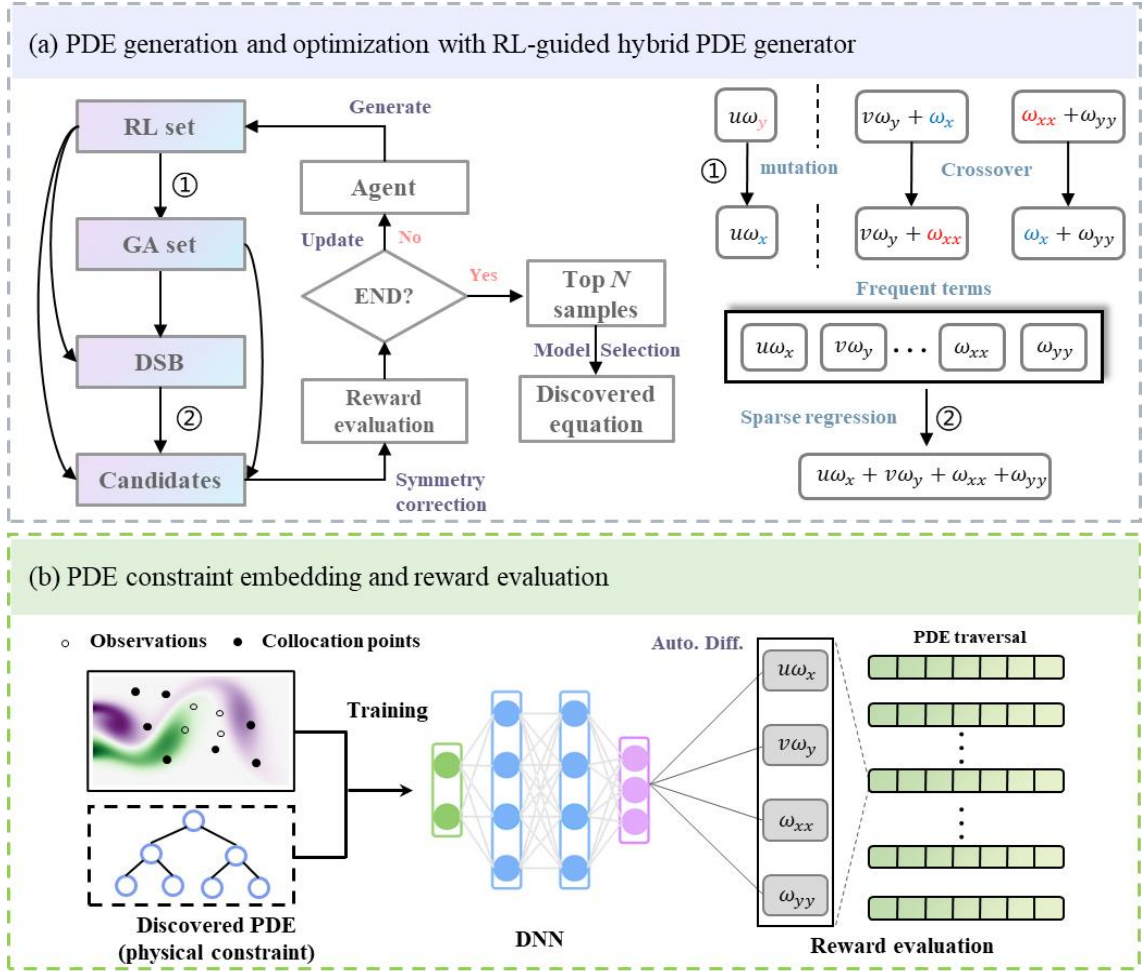

**FIG. 2**. Schematic diagram of: (a) PDE generation and optimization with a RL-guided PDE generator; and (b) PDE constraint embedding and reward evaluation.

## 3. Reward evaluation

After generating PDE expressions, it's necessary to further evaluate their rewards. Here, we first need to split the PDE expressions into function terms according to '+' or '-' on the top of trees. Then, based on Sequential Threshold Ridge regression (STRidge) [2], we determine the coefficients of function terms as follows:

$$\hat{\xi} = \arg\min_{\xi} |\Theta(u,x)\xi - u_t|_2^2 + \kappa \|\xi\|_2^2 \tag{5}$$

When $\kappa = 0$, the regularization term becomes ineffective and this method reduces to the Sequentially Thresholded Least Squares method (STLS). With the function terms and their coefficients determined, a complete PDE representation can be obtained. We then evaluate the rewards of the generated PDE expressions with the following metrics:

$$R(\tau) = \frac{1 - \zeta_1 \times d_{max} - \zeta_2 \times n}{1 + RMSE}, RMSE = \sqrt{\frac{1}{N_c} \sum_{i=1}^{N_d} (\Theta(u_i, x_i)\xi - u_{t_i})^2} \tag{6}$$

where $d_{max}$ represents the maximum depth of the subtree structure corresponding to the function terms, and $n$ denotes the number of function terms. The former is used to measure the complexity of a single equation term, while the latter is used to measure the complexity of the PDE traversal. $\zeta_1$ and $\zeta_2$ represent the penalties for the above two complexities, used to ensure the simplicity of the equation form. The RMSE in the numerator is used to measure the fitness of observations.

Notably, compared with common numerical differentiation and interpolation methods utilized in SINDy series work [2,12,34], we utilize the neural network as the reward evaluator to evaluate the generated PDE samples. Specifically, a NN-based predictive model is built to fit the system response $u$ and smooth noisy data. Then, the NN can be utilized to make predictions and evaluate the partial derivatives based on automatic differentiation in a mesh-free manner. The concrete process is illustrated in FIG. 2 (b).

## 4. Optimization with the risk-seeking policy-gradient method

After reward evaluation, the next step is to instruct the agent to optimize the generated expressions. In this problem, the parameters of the agent cannot be updated directly by the reward function using the common gradient descent method, therefore the reinforcement-learning strategy is utilized. In detail, we adopt the risk-seeking policy gradient method, which is dedicated to improving the performance of the better-fitting expressions in the generated samples [28]. This is mainly because the PDE discovery task aims to uncover the best-performing equation representation rather than to maximize the expectation of all of the samples' rewards. We sort the rewards of all samples and select the $(1-\varepsilon)$-quantile reward $\tilde{q}_\varepsilon(R)$ as a threshold and samples with rewards higher than $\tilde{q}_\varepsilon(R)$ can be utilized as training samples. The goal of optimization can be represented as

$$J_{\text{risk}}(\theta;\varepsilon) \doteq \mathrm{E}_{\tau \sim p(\tau|\theta)}\left[R(\tau) \mid R(\tau) \geq R_\varepsilon(\theta)\right] \tag{7}$$

Its gradient can be calculated using the following equation.

$$\nabla_\theta J_{\text{risk}}(\theta;\varepsilon) \approx \frac{\lambda_{pg}}{\varepsilon N_{train}} \sum_{i=1}^{N_{train}} \left[R(\tau^{(i)}) - \tilde{q}_\varepsilon(R)\right] \cdot \mathbf{1}_{R(\tau^{(i)}) \geq \tilde{q}_\varepsilon(R)} \nabla_\theta \log p(\tau^{(i)} \mid \theta) \tag{8}$$

where $N$ refers to all of the PDE expressions generated from the hybrid generator at each iteration; $\lambda_{pg}$ refers to a hyperparameter that measures the importance of the risk-seeking policy-gradient method; and $\mathbf{1}_x$ denotes a conditional judgment that returns 1 when the condition represented by $x$ is true and 0 otherwise. More details can be found in the DISCOVER[10].

## 5. Model selection by considering the stability of coefficients

Through iterative optimization, the hybrid generator produces and accumulates some high-quality PDE samples. We perform a preliminary evaluation of them through the elaborated reward function. Note that noisy and sparse data inevitably leads to the introduction of bias by the reward evaluator when assessing each generated equation. Consequently, there may exist some generated equations that are not the true equation but possess higher rewards. To alleviate the impact of noise and maintain the accuracy of the evaluation while minimizing time costs, we first retain the top $K$ candidates during the search process. The evaluation is then further refined from the perspective of balancing fitting accuracy and coefficient stability. In related studies, the statistical performance of coefficients on a subset of full data has also been taken into consideration like E-SINDy [35] and other methods based on the moving horizon technique [36,37]. In this paper, we propose a comprehensive evaluation metric to comprehensively assess the discovered PDEs. The best PDEs are selected through majority voting, as shown in FIG. 3.

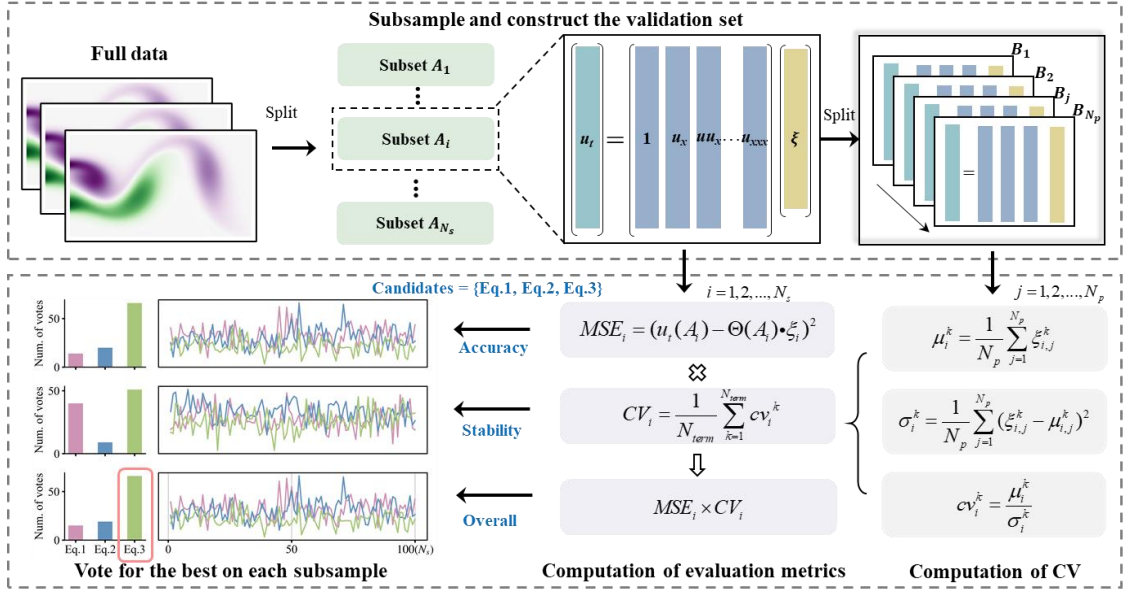

**FIG. 3.** Schematic of model selection based on data fitting and coefficients stability.

The primary idea is that distinguishing between necessary and redundant terms through a single fit on the full dataset is challenging. The overfitting problem becomes worse when measurements are noisy and sparse. However, it is expected that essential function terms will consistently possess non-zero coefficients when a series of fits are conducted on various sub-sampled datasets. In contrast, the redundant terms, which primarily serve as fitting for noise, are likely to have a drastic change of coefficients and fitting accuracy on each subsample. Taking one of the PDE candidates with a function library $\Theta$ as an example, we first utilize randomly sampled collocation points and deep neural network (DNN) to generate the dataset $U_c = \{\hat{u}(t_i, x_i)\}_{i=1}^{N_c} \subseteq (0,T] \times \Omega$, and $N_s$ bootstrapped subsets are sampled with equal sizes of $N_c / 2$. The coefficients of each PDE candidate are then calculated based on the regular least squares method and we further evaluate the MSE between the LHS and the RHS of the PDE as a measure of the accuracy of data fitting. It is represented by

$$\text{MSE}_i = (u_t(A_i) - \Theta(A_i) \cdot \xi_i)^2, \quad i=1,2,...,N_s \tag{9}$$

where $A_i$ refers to the subset of full data. Then, subsets are further split into subsamples $B_j \subset \{1,2,...,N_c/2\}$, $j=1,2,...,N_p$ of equal size $N_c/4$. For each subsample in the subset $A_i$, the coefficients are recalculated, and the coefficients of variation (cv) for the term $k$ in $\Theta$ can be obtained by

$$cv_i^k = \frac{\mu_i^k}{\sigma_i^k} \tag{10}$$

where $\mu_i^k$ and $\sigma_i^k$ refer to the mean and standard deviation of the $k^{th}$ term of the current PDE candidate in subset $A_i$, respectively. For the PDE candidate as a whole, CV can be further evaluated by

$$CV_i = \frac{1}{N_{term}} \sum_{k=1}^{N_{term}} cv_i^k \tag{11}$$

where $N_{term}$ refers to the size of the function library. We can use the product of the MSE and CV as a measure of the performance of each PDE candidate on each subset. Finally, we aggregate and rank these measures and take a majority vote to select the best PDE. The detailed procedures are illustrated in FIG. 3.

### C. Automatic physics embedding process

When applied to sparse and noisy data, common neural networks possess some desired features that enable them to exert smoothing and denoising effects to some extent. However, overparameterization or extremely poor data quality may lead these models to overfit the noise, resulting in the introduction of significant biases in the predictions. The practice of embedding physics, especially governing equations, into neural networks named physics-informed neural networks has been widely studied and discussed [21,38–40]. It has been proven to be an effective way to improve the prediction accuracy and robustness to noise. In this section, we will demonstrate how R-DISCOVER embeds discovered equations into neural networks as physical constraints automatically by means of structural properties of the PDE tree. Without halting the training midway, the framework achieves a complete and complementary closed-loop process of knowledge discovery and knowledge embedding.

#### 1. Automatic residual evaluation

After the discovering process, an initially identified PDE can be obtained, and while it may not be entirely accurate, the extracted equation terms still preserve valuable physical information. According to the PINN-training procedures, the governing equation is incorporated into the loss function to constrain the output of neural networks. Hence, the key to imposing physical constraints is to solve for the residuals on the RHS and LHS of the identified PDE. However, the exact form of PDEs cannot be predetermined at each iteration, and which means we cannot explicitly write them into the program as is typically done with common PINN does. In this paper, we adopt an approach from automatic machine learning, specifically an automatic knowledge embedding framework named AutoKE[41], which automatically constructs computational graphs by means of traversing PDE trees. We first reconstruct the PDE traversal obtained from the discovering process into a PDE tree. Since all of the operations are performed using operators from deep learning frameworks, such as Pytorch[42], the computational graph can be constructed automatically during the traversal of the PDE tree to calculate the PDE residual. Note that available prior information, for example, the boundary or initial conditions, can also be incorporated automatically into the current framework as extra physical constraints. Partial derivatives can be evaluated by the automatic differentiation in the neural network, as shown in FIG. 2 (b).

## 2. Physics embedding with identified PDEs and observations incorporated

To ensure the physical and data consistency when constructing a predictive model in the PINN-training paradigm, two aspects of loss need to be considered: data loss and physics loss. They can be represented as follows:

$$L_d(W) = \frac{1}{N_m} \sum_{i=1}^{N_m} |\hat{u}_i - NN_W(x_i, t_i)|^2 \tag{12}$$

$$L_p(W) = \frac{1}{N_c} \sum_{i=1}^{N_c} |R(W, \tau^*; U_c)|^2 = \frac{1}{N_c} \sum_{i=1}^{N_c} |u_t(x_i, t_i) - \Theta(u_i, x_i) \cdot \xi|^2 \tag{13}$$

where $W$ represents the trainable parameters in DNN and $\tau^*$ is the initially identified PDE obtained by model selection. The residual has been evaluated by traversing the PDE tree, whose coefficients are set to be non-trainable until the last iteration.

Since the noisy data may increase the bias of the network prediction, a local sampling strategy is proposed as an optional smoothing trick in the training process. We resample supplementary collocation points around the observations and create a new loss term by

$$L_l(W) = \frac{1}{N_{sc}} \sum_{i=1}^{N_{sc}} |u_t(x_i, t_i) - \Theta(u_i, x_i) \cdot \xi|^2 \tag{14}$$

where $N_{sc}$ is the number of collocation points resampled. Centered on each observation point, a rectangular sampling domain can be constructed with side lengths $\Delta t_{min}$ and $\Delta x_{min}$. $\Delta t_{min}$ and $\Delta x_{min}$ denote the minimum intervals for observational data along temporal and spatial dimensions respectively. The default volume of data to be resampled is set to $N_{sc} = 10 \times N_m$. Consequently, the total PINN loss can be represented as

$$L(W) = L_d + \lambda_1 L_p + \lambda_2 L_l \tag{15}$$

where $\lambda_1$ and $\lambda_2$ represent the hyperparameters for measuring the importance of corresponding loss terms. The default value of the latter $\lambda_2$ is 0. However, according to experience, when encountering highly noisy and sparse observations, this loss term can have a smoothing and corrective effect on the prediction.

Automatically embedding the mined physical information into the neural network can effectively enhance the robustness of the prediction. This improvement is beneficial for predicting state variables and evaluating partial derivatives in the discovery process. Note that scarce data and extremely high noise may result in significant discrepancies between the identified equations and correct equations, which may introduce relatively large biases. Therefore, we utilize the above model selection method to ensure that the correct equation terms are identified in the optimal equation. After iteratively alternating between the processes of discovery and embedding, the coefficients in the discovered equations become closer to the true values. The reward evaluator, which incorporates the discovered physical constraints, can be rectified iteratively and make more accurate predictions.

## III. RESULTS

### A. Data descriptions and evaluation metrics

In this section, we verify the framework's ability to identify governing equations of six nonlinear dynamic systems in the presence of limited and highly noisy data. These six equations mainly include the classical Burgers' equation, Fisher–Kolmogorov–Petrovsky–Piskunov (Fisher–KPP) equation and nonlinear Fisher-KPP equations with a square of the spatial derivative, the Kuramoto–Sivashinsky (KS) equation with high-order spatial derivatives, Navier-Stokes equations with multiple state variables and 3D Gray–Scott reaction-diffusion model with 2 PDEs included and only low-resolution data available. The descriptions of these six equations are shown in Table II.

The total training process includes three parts, pretraining with the observations, discovering open-form PDEs, and embedding PDE constraints into the DNN. The latter two are alternately updated as illustrated in Sec. II. In the experiments, to minimize the training time, we carry out only two rounds of the alternating iteration process. During the pretraining parts, we divide the observations into training sets and validation sets at a ratio of 8:2. We use the Adam optimizer to fit the noisy data and determine the end of training based on the performance of the validation set. During the discovering process with symbolic representation, only the top 10% of candidate expressions are selected to update the agent. More detailed descriptions of the hyperparameters are provided in the supplementary material.

We use four metrics to evaluate the ability of the proposed framework to identify governing equations. The first one is the relative error of the coefficients, which is used to measure the difference between the uncovered coefficients and the true coefficients, as follows

$$E = \frac{1}{N} \sum_{i=1}^{N} \frac{|\xi_i^* - \xi_i|}{|\xi_i|} \times 100\% \tag{17}$$

where $N$ denotes the number of function terms; $\xi_i$ and $\xi_i^*$ represent the estimated and true values of the $i^{th}$ term, respectively. This metric can only be applied when the discovered function terms align with the true equations. In addition, we introduce two other metrics, relative coefficient error $E_2$ and the true positive rate (TPR), to measure the performance when the mined equations are not exactly correct. They are evaluated by

$$E_2 = \frac{\|\xi^* - \xi\|_2}{\|\xi\|_2}, \quad TPR = \frac{TP}{TP + FN + FP} \tag{18}$$

True positive (TP) denotes the count of non-zero coefficients in ξ that have been accurately identified while false negative (FN) refers to the number of coefficients in ξ that are incorrectly identified as zero. On the other hand, false positive (FP) corresponds to the number of coefficients in ξ that are mistakenly identified as non-zero. Higher values of true positive rate (TPR) indicate that the actual equation form has been approximated more accurately, and a TPR value of 1 signifies that the correct equation form has been successfully retrieved.

**TABLE II.** Descriptions of canonical dynamic systems utilized in the experiments.

| Equation | | Descriptions |
|---|---|---|
| Burgers' equation[43,44] $u_t = -uu_x + 0.1u_{xx}$ | 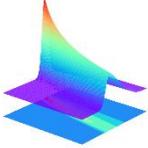 | $x \in [-8,8]_{nx=256}, t \in [0,10]_{nt=101}$, Sample ratio 3.19%~19.34%, Noise level ($\sigma$): 10%~125% |
| Fisher-KPP equation[45,46] $u_t = 0.02u_{xx} + 10u - 10u^2$ | 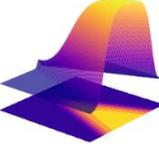 | $x \in (-1,1)_{nx=199}, t \in (0,1)_{nt=99}$, Sample ratio: 10%~25%, Noise level ($\sigma$): 10%~100% |
| Nonlinear Fisher-KPP[47,48] $u_t = 0.02uu_{xx} + 0.02u_x^2 + 10u - 10u^2$ | 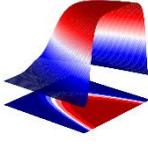 | $x \in (-1,1)_{nx=199}, t \in (0,1)_{nt=99}$, Sample ratio 25%~50%, Noise level ($\sigma$): 10%~50% |
| Kuramoto-Sivashinsky[49,50] $u_t = -uu_x - u_{xx} - u_{xxxx}$ | 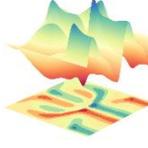 | $x \in [-10,10]_{nx=512}, t \in [0,50]_{nt=256}$, Sample ratio: 5%~50%, Noise level ($\sigma$): 10%~30% |
| Navier-Stokes equation[51,52] $\omega_t = 0.01\omega_{xx} + 0.01\omega_{yy} - u\omega_x - v\omega_y$ | 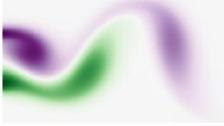 | $x \in [0,6.5]_{nx=325}, y \in [0,3.4]_{ny=170}$, $t \in [0,30]_{nt=150}$, Sample ratio: 0.72%, Noise level ($\sigma$): 5%~25% |
| 3D Gray–Scott reaction-diffusion model[53,54] $\begin{cases} u_t = 0.02\nabla^2 u - uv^2 + 0.014(1-u) \\ v_t = 0.01\nabla^2 v + uv^2 - 0.067v \end{cases}$ | 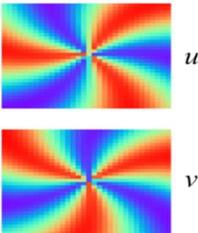 | $x \in [-1.25,1.25)_{nx=32}, y \in [-1.25,1.25)_{ny=32}$, $z \in [-1.25,1.25)_{nz=32}, t \in [0,10]_{nt=100}$, Sample ratio: 10%, Noise level ($\sigma$): 0~10% |

Given the spatio-temporal information, the neural network trained in this framework can be used to further predict the physical field of state variables of interest, which is the advantage of combining prediction network (DNN) and PINN compared with other numerical discovery methods, like SINDy. Therefore, another evaluation metric $L_2$ error is utilized to measure the relative error between the reconstructed physical field and the real physical field.

$$L_2(u_{pred}, u_{true}) = \frac{\|u_{pred} - u_{true}\|_2}{\|u_{true}\|_2} \tag{19}$$

We added Gaussian noise to the observations to verify the robustness of the framework. The way of applying the noise is shown below

$$u(x,t) = u(x,t) + \sigma \cdot std(u) \cdot \mathrm{N}(0,1) \tag{20}$$

## B. Experiments

### 1. Discovering Burgers' equation

The Burgers' equation is a prototypical example of a nonlinear partial differential equation, which exhibits a wide range of complex phenomena, including shock waves, rarefaction waves, and turbulence [43,44]. The equation considered in this paper takes the form of $u_t = vu_{xx} - uu_x$, where $u$ denotes the one-dimensional velocity field, and $v$ represents the kinematic viscosity. The ground truth is obtained through the finite difference method. We provide the concrete performance when discovering Burgers' equation under different noise and data volumes. As shown in FIG. 4, the coefficients of the discovered equations diverge from true values as the noise level increases and the data quantity decreases. This divergence continues until incorrect function terms are mined or certain true function terms are omitted. When the correct equation terms are identified, the physical constraints can help improve the accuracy of the physical field derived from the reconstruction. Conversely, it can also increase the error. With more than 2000 data points (about 8% of the total data points), the proposed framework can successfully mine the correct equations even in the presence of up to 125% white noise.

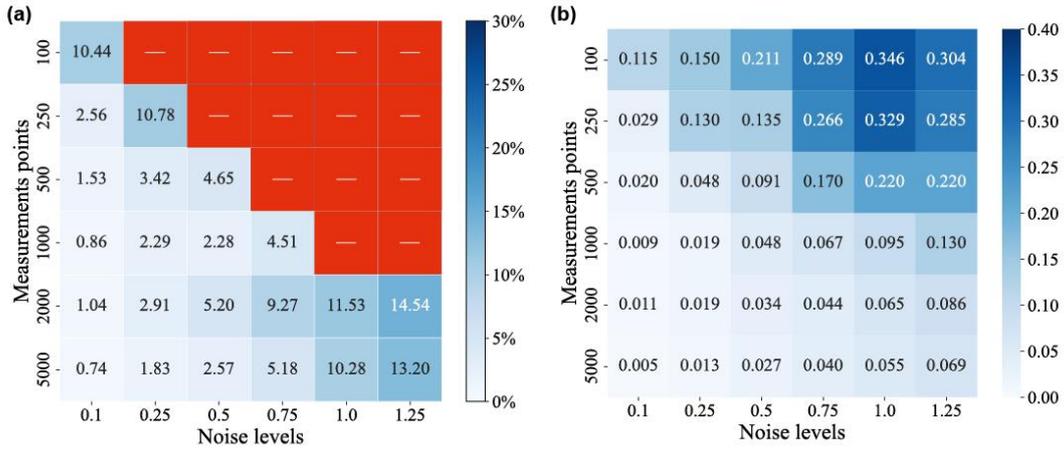

**FIG. 4.** Mean of relative coefficient $E$ (a) and the relative full-field $L_2$ error (b) under different fractions of data and white noise. The red square indicates that the discovered equation terms are not consistent with the ground truth.

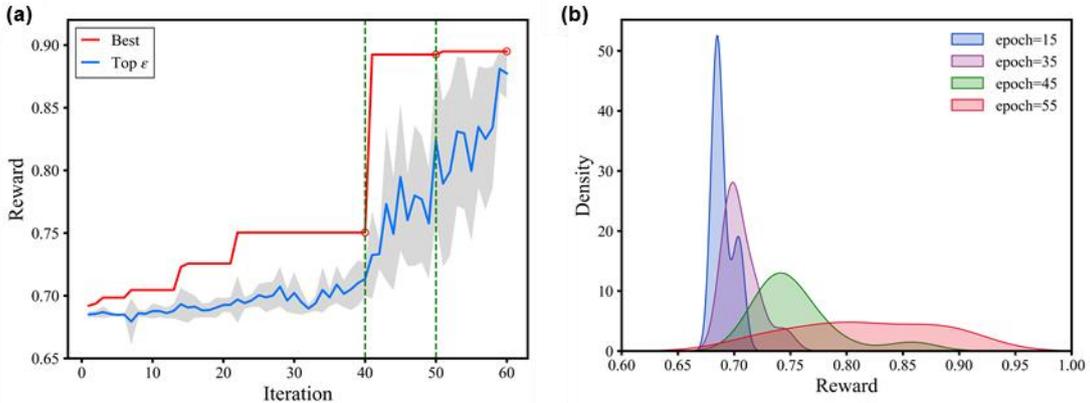

**FIG. 5.** (a): the evolution of the best reward and top $\varepsilon$ fraction of rewards for 5000 data points (20% of the total data points) with 100% Gaussian noise. The positions marked by the green dashed line and the red circle indicates the start of each new round of the search process after the DNN is updated. (b) The Gaussian kernel density estimate of the top fraction of rewards.

FIG. 5 illustrates the optimization process in the presence of a data volume of 5000 data points and 100% noise. The robustness of the prediction results can be enhanced by constraining the output of the DNN with the discovered equations after each alternate iteration of the process, leading to a notable increase in reward. The final identified equation is as follows:

$$u_t = -0.958uu_x + 0.0836u_{xx} \tag{21}$$

FIG. 6 demonstrates the predicted results by DNN when the discovered PDE constraint is incorporated. It can be seen that the predicted results are consistent with the exact solution, which indicates the R-DISCOVER can discover the Burgers' equation directly from the noisy observations.

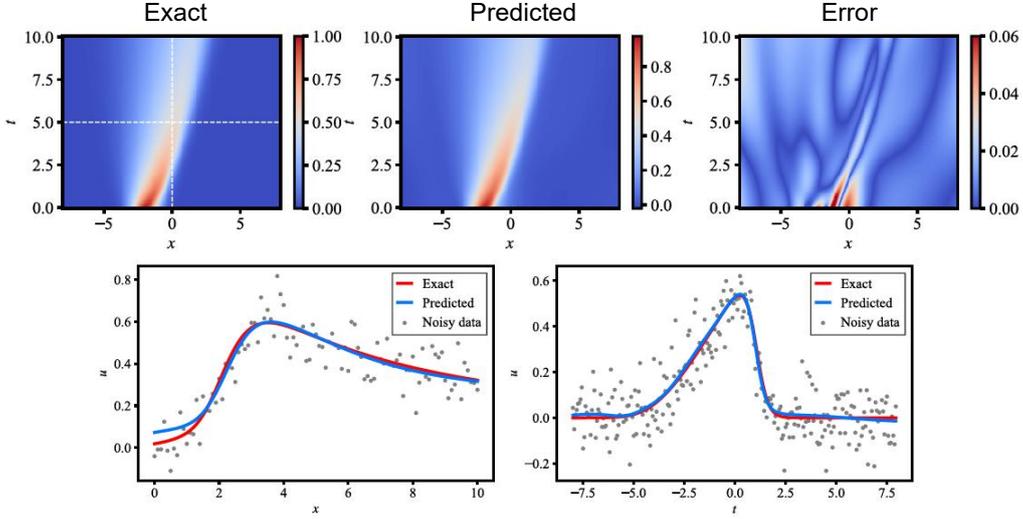

**FIG. 6.** The comparison between the true and predicted solution. The white dashed line refers to the location of the line plot below.

## 2. Discovering the biological transport model described by Fisher-KPP equations

The Fisher-KPP equation is a simplified version of the reaction-diffusion equation. The equation has been the subject of extensive study in the fields of mathematical biology and is used to model a wide range of physical systems, including combustion [46], epidemics [55], and tumor growth [45]. In this experiment, we consider a one-dimensional classical Fisher-KPP equation taking the form of $u_t = Du_{xx} - ru/k - ru^2$, where $u$ represents the density of the population; $D$ denotes the diffusion coefficient; $r$ is the intrinsic growth rate; and $k$ is the carrying capacity of the environment. We also introduce a nonlinear Fisher-KPP equation taking the form of $u_t = Duu_{xx} + Du_x^2 - ru/k - ru^2$, where an extra square of the spatial derivative $(u_x)^2$ cannot be identified by most overcomplete library methods in the default library settings. In the absence of a priori knowledge, assuming that the governing equations are composed of monomials of $u$ and its partial derivatives with a more general form of $u^m(\partial_x^n u)^k$. The number of total possible terms in the overcomplete library can be represented with $N_l(m,n,k) = (m+1)(n*k+1)$, which grows dramatically when large orders are considered. For example, for $m = n = 3$ and $k = 2$, the size of the library is $N_l(3,3,2) = 28$, but $m = n = 5$ and $k = 3$, it will become $N_l(5,5,3) = 96$. This can greatly increase the DNN training time and computational burden, and the size will grow exponentially for high-dimensional dynamic systems with multiple state and spatial variables.

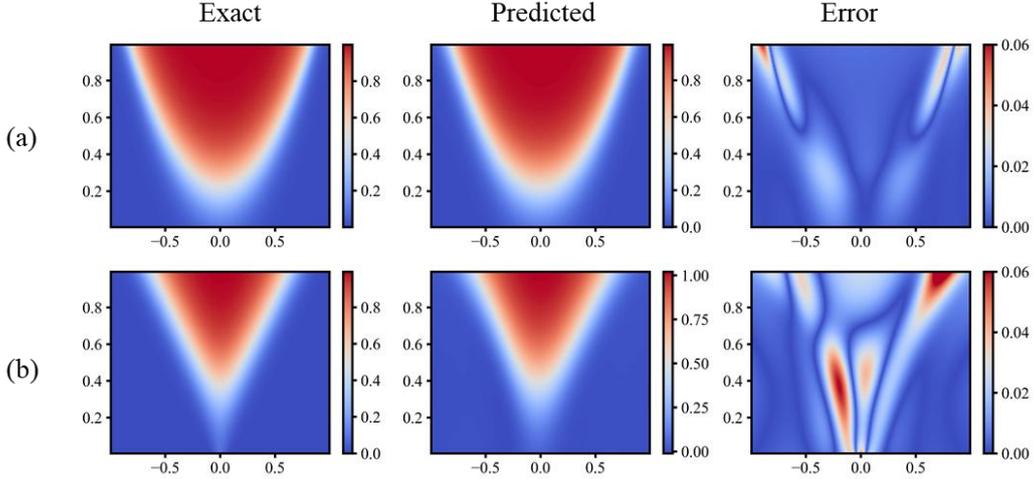

**FIG. 7.** Predicted solutions of (a) classical with 100% noise added and 5000 data points available; (b) nonlinear Fisher-KPP equations with 50% noise added and 5000 data points available.

Since our framework only considers function terms of the optimal expression when embedding PDEs, the cost of storage and optimization can be effectively alleviated. When mining classical Fisher-KPP as shown in Table III, our framework can correctly identify the correct equation terms under at most 100% noise added and 25% of total data available. For the Non-linear Fisher-KPP equation, due to the stronger nonlinearity of the system, the evaluation of function terms will be more complicated and challenging. When uncovering this equation, the framework can uncover all of the correct function terms under at most 50% of the noise, but the error of equation coefficients is relatively larger compared to the classical one. The details are demonstrated in Table IV. The predicted solutions of those two equations with maximum noise are presented in FIG. 7. In high-noise scenarios, while the coefficients of R-DISCOVER may deviate slightly from the actual results, the form of the equations remains mostly accurate. This experiment demonstrates the flexibility of R-DISCOVER in discovering various nonlinear systems and its robustness in extracting common equation information from noisy data.

**TABLE III.** Discovery of classical Fisher-KPP equation under different levels of noise and data volume.

| | | Correct PDE: $u_t = 0.02u_{xx} + 10u - 10u^2$ | | |
|---|---|---|---|---|
| Noise level | Data Volume | Identified PDE | $E$ (%) | $L_2$ error |
| 50% | 10,000 | $u_t = 0.019u_{xx} + 9.37u - 9.939u^2$ | 2.08 | 0.016 |
| 75% | (50% of total data) | $u_t = 0.021u_{xx} + 9.520u - 9.593u^2$ | 4.62 | 0.029 |
| 100% | | $u_t = 0.021u_{xx} + 9.467u - 9.520u^2$ | 5.04 | 0.031 |
| 50% | 5000 (25% of total data) | $u_t = 0.022u_{xx} + 9.675u - 9.539u^2$ | 5.95 | 0.026 |
| 75% | | $u_t = 0.020u_{xx} + 9.157u - 8.974u^2$ | 6.23 | 0.040 |
| 100% | | $u_t = 0.022u_{xx} + 9.141u - 8.890u^2$ | 9.89 | 0.049 |

**TABLE IV.** Discovery of nonlinear Fisher-KPP equation under different levels of noise and data volume.

| | | Correct PDE: $u_t = 0.02uu_{xx} + 0.02u_x^2 + 10u - 10u^2$ | | |
|---|---|---|---|---|
| Noise level | Data Volume | Identified PDE | $E$ (%) | $L_2$ error |
| 10% | 10,000 (50% of total data) | $u_t = 0.019uu_{xx} + 0.019u_x^2 + 9.994u - 10.011u^2$ | 2.54 | 0.006 |
| 25% | | $u_t = 0.015uu_{xx} + 0.015u_x^2 + 9.979u - 9.963u^2$ | 12.65 | 0.010 |
| 50% | | $u_t = 0.010uu_{xx} + 0.010u_x^2 + 9.710u - 9.780u^2$ | 26.28 | 0.019 |
| 10% | 5000 (25% of total data) | $u_t = 0.019uu_{xx} + 0.019u_x^2 + 9.945u - 9.936u^2$ | 2.80 | 0.007 |
| 25% | | $u_t = 0.014uu_{xx} + 0.014u_x^2 + 9.979u - 9.963u^2$ | 15.12 | 0.013 |
| 50% | | $u_t = -0.021u_x + 0.001u_x^4 + 9.840u - 9.896u^2$ | \ | 0.020 |

### 3. Discovering Kuramoto–Sivashinsky equation

The KS equation utilized here describes the evolution of a one-dimensional fluid flow in the presence of nonlinearity, dispersion, and diffusion[49,50]. The term $uu_x$ represents the advection term, the term $u_{xx}$ represents the diffusion term, and the term $u_{xxxx}$ represents the fourth-order dispersion term. Compared to other equations, higher-order derivatives increase the difficulty of discovery of the equation under noisy data. To cope with this approach, we introduce the weak-form method in the evaluation of the equation terms, which can reduce the order of the derivatives and improve the accuracy of the derivative evaluation. We take the collocation points uniformly in the defined computational region $x \in [-6,6]$ and $t \in [10,40]$, where 1200 points and 300 points are uniformly taken in the space and time direction respectively.

**TABLE V.** Discovery of KS equation for 26215 data points (20% of total measurements) with different levels of noise.

| | Correct PDE: $u_t = -uu_x - u_{xx} - u_{xxxx}$ | | |
|---|---|---|---|
| Noise level | Identified PDE | Error (%) | $L_2$ error |
| 5% | $u_t = -0.996uu_x - 0.989u_{xx} - 0.988u_{xxxx}$ | 1.01 | 0.009 |
| 10% | $u_t = -0.982uu_x - 0.984u_{xx} - 0.988u_{xxxx}$ | 1.14 | 0.016 |
| 20% | $u_t = -0.960uu_x - 0.942u_{xx} - 0.938u_{xxxx}$ | 4.95 | 0.020 |
| 30% | $u_t = -0.941uu_x - 0.885u_{xx} - 0.879u_{xxxx}$ | 9.46 | 0.029 |

TABLE VI. Discovery of KS equation with 15% noise and different volumes of data.

Correct PDE: $u_t = -uu_x - u_{xx} - u_{xxxx}$

| Data Volume | Identified PDE | Error (%) | $L_2$ error |
|---|---|---|---|
| $N_m$=52,429 (40%) | $u_t = -0.988uu_x - 0.986u_{xx} - 0.988u_{xxxx}$ | 0.90 | 0.013 |
| $N_m$=39,322 (30%) | $u_t = -0.982uu_x - 0.984u_{xx} - 0.988u_{xxxx}$ | 1.15 | 0.016 |
| $N_m$=26,214 (20%) | $u_t = -0.975uu_x - 0.954u_{xx} - 0.953u_{xxxx}$ | 3.55 | 0.016 |
| $N_m$=13,108 (10%) | $u_t = -0.978uu_x - 0.953u_{xx} - 0.948u_{xxxx}$ | 3.65 | 0.021 |
| $N_m$=6,554 (5%) | $u_t = -0.931uu_x - 0.896u_{xx} - 0.903u_{xxxx}$ | 8.64 | 0.032 |

In this case, we tested the robustness of the framework with different data volumes and noise levels. As shown in Table V, the correct equations can be identified with only 20% of the data from the original simulation points are used when up to 30% noise is added. More detailed results about the sensitivity of data volume are shown in Table VI. This example demonstrates that our framework can effectively identify the governing equations of complex dynamic systems with high-order derivatives.

## 4. Discovering Navier-Stokes equation

The Navier-Stokes equations are widely used in computational fluid dynamics to analyze and predict the behavior of fluids in various situations [51,52]. Notably, the introduction of the multiple state variable ($\omega, u, v$) and two-dimensional space ($x, y$) increases the number of basic operators. It significantly increases the number of possible combinations of symbols, which in turn expands the search space and increases the difficulty of discovering the correct equation. On the other hand, the correct equation contains four terms with different derivatives, which further increases the challenge of using a sequence model to generate the equation representation. As shown in FIG. 8, under the condition of 25% noise, we compare the RL generator (only the LSTM agent utilized) with the hybrid generator in discovering the equations. It can be found that the common RL method is unable to find the correct combination of equations in a shorter period, and tends to be trapped in a local optimal. The wrong equations will further mislead the training of DNN. For the hybrid generator, it is possible to collect the possible function terms by fully utilizing the compact subtree library, and based on the correction of the symmetry of the space, the correct equations can be found in a shorter iteration step.

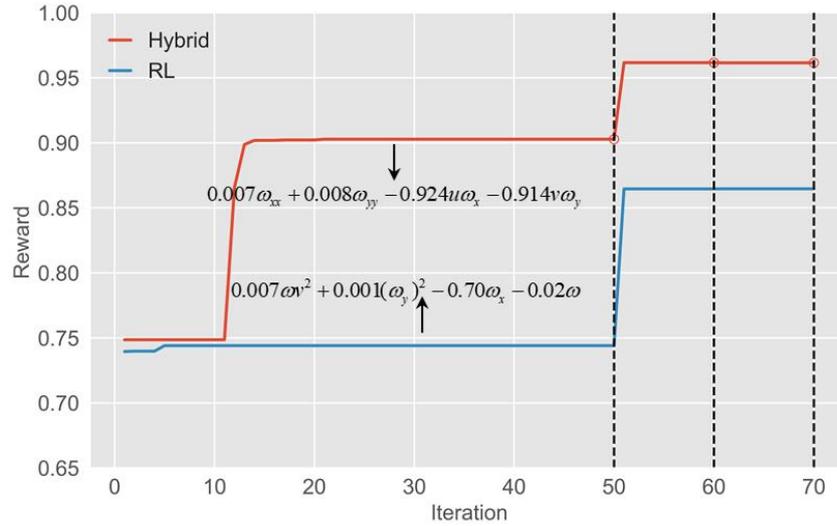

**FIG. 8.** The evolution of the best reward with RL generator and hybrid generator.

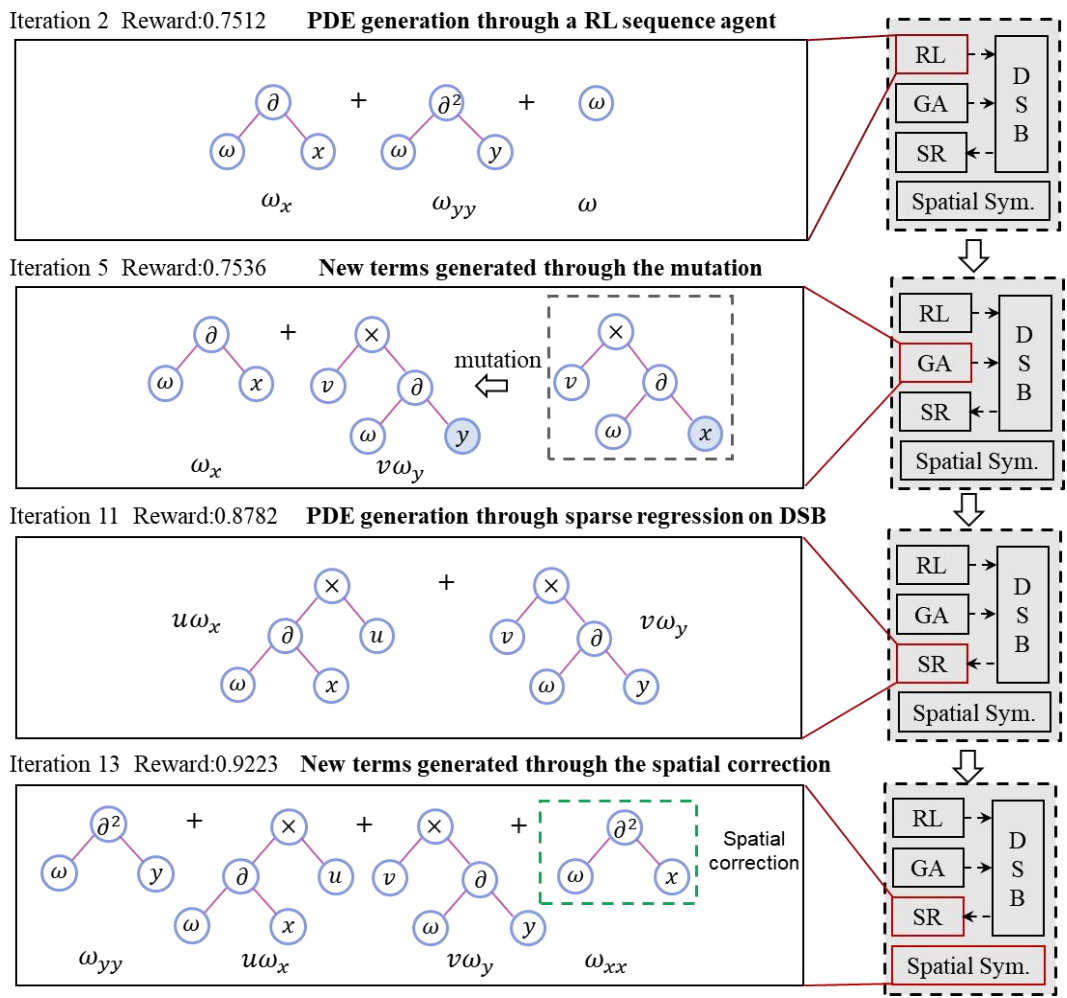

**FIG. 9.** Evolution of discovering NS equation.

**TABLE VII.** Discovery of NS equation under different noise levels.

Correct PDE: $\omega_t = 0.01\omega_{xx} + 0.01\omega_{yy} - u\omega_x - v\omega_y$

| Noise level | Identified PDE | Error (%) | $L_2$ error |
|---|---|---|---|
| 10% | $\omega_t = 0.0094\omega_{xx} + 0.0094\omega_{yy} - 1.011u\omega_x - 0.994v\omega_y$ | 2.88 | 0.082 |
| 15% | $\omega_t = 0.0087\omega_{xx} + 0.0097\omega_{yy} - 0.987u\omega_x - 0.99v\omega_y$ | 3.93 | 0.083 |
| 25% | $\omega_t = 0.0084\omega_{xx} + 0.0097\omega_{yy} - 1.00u\omega_x - 0.982v\omega_y$ | 5.20 | 0.097 |

FIG. 9 further illustrates the evolution of function terms during the optimization process and highlights the significance of a hybrid PDE generator, where RL generates expressions as initial samples, GA increases the stochasticity of the expressions, and DSB efficiently generates the final equations based on the possible combinations of the function terms. At the same time, spatial correction can further help to produce final physically correct equations. Table VII demonstrates the performance of the framework to uncover the correct equations under different noise levels.

### 5. Discovering 3-D Gray–Scott reaction-diffusion model

Gray–Scott reaction–diffusion model is a two-component reaction-diffusion system, often used to model pattern formation and self-organization in a variety of scientific contexts [53,54]. The main difficulty in discovering this equation from data is that the dynamic system is governed by two PDEs and the available observations are both high-dimensional and of low resolution. This results in two key issues. First, the symbolic library, which is comprised of two state variables and four spatio-temporal variables, leads to an expansive search space and an increased susceptibility to local optimal solutions. Second, accurately solving the partial differentials becomes increasingly challenging with low-resolution data.

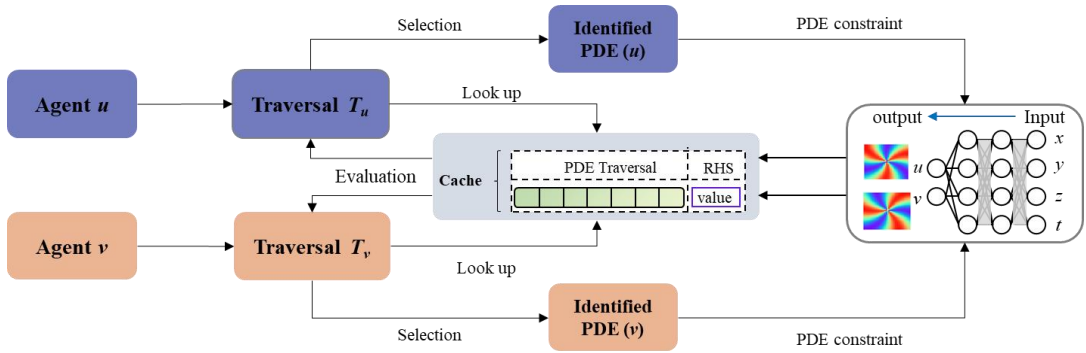

**FIG. 10.** Framework for discovering 3D Gray–Scott reaction–diffusion model.

To address these issues, we propose a modified discovering architecture, as depicted in FIG. 10. For each equation to be determined, a different agent is employed to generate the corresponding PDE traversal. Since the primary time bottleneck lies in the evaluation of the generated traversal. Specifically, in the computation of the RHS of the equation, we adopt a caching mechanism. An extensible dictionary is constructed to record the traversal and its corresponding RHS value. This approach mitigates the need to repeat computations for generated traversals, thereby enhancing computation efficiency.

For the derivative evaluation, we use a PDE-constrained neural network to generate more metadata to avoid the problem of inaccurate prediction caused by low-resolution data. With 20% noise and 20% measurements of the total data volume, our framework is still able to find the final accurate equation with a 2.96% relative coefficient error, and the results are shown below:

$$\begin{cases} u_t = 0.019 u_{xx} + 0.020 u_{yy} + 0.018 u_{zz} - 0.975 uv^2 - 0.019 u + 0.014 \\ v_t = 0.010 v_{xx} + 0.010 v_{yy} + 0.009 v_{zz} + 1.020 uv^2 - 0.069 v \end{cases} \quad (22)$$

## C. Comparison with other related methods

Embedding the uncovered physical equation into the predictive model (DNN) can significantly enhance the robustness of our framework to both limited and noisy data. This idea has been incorporated in many studies on PDE discovery, some of which explore different structures of neural networks [23–25], while others employ more flexible methods for equation representations [27]. Some methods focus on data preprocessing [26] and model selection [25,36]. However, to the best of our knowledge, few studies have compared the performance and parameter robustness of these methods under different noise levels. In this study, we use DeepMod [23] and PINN-SR [24] as baseline models for a detailed comparison and discussion. These two models are considered to be the most representative and influential methods among the discovery methods based on PINN-type training. Both methods construct a compact candidate library using prior knowledge, and add the residuals of the linear combinations of all the function terms and the LHS of the equation $u_t$ as a constraint into the training process of the neural network. $L_1$ regularization is also incorporated to ensure sparsity. The difference mainly lies in the determination of the final parsimonious PDE representation. The former achieves this by normalizing the function terms in the library and using the statistical properties of coefficients as the hard threshold to filter out trivial terms with small coefficients. PINN-SR, on the other hand, adopts an alternating direction optimization strategy between DNN training and the STRidge method. Note that while the compact and fixed library is necessary for efficient optimization, it limits the capacity to discover complex combinations, such as fractional structure and compound terms [9].

Compared to the other two methods, R-DISCOVER, based on previous work DISCOVER, utilized symbolic representation to present PDEs, which enabled it to discover open-form PDEs even without prior knowledge. Meanwhile, its ability to handle noisy and sparse data is significantly enhanced. We compare the performance of PINN-SR, DeepMod, and DISCOVER with our proposed framework on Burgers' equation, nonlinear Fisher-KPP equation, and KS equation. To enable a fair comparison, both DeepMod and PINN-SR were configured with basis function libraries of identical size, carefully incorporating all terms associated with the target equations. Related hyperparameters were kept consistent with those specified in the originating article and not adjusted under different noise levels.

**TABLE VIII.** Comparison of different PDE discovery methods. Two levels of noise are considered. Low noise refers to 0.1, 0.1, and 0.1, and high noise refers to 0.5, 0.3, and 0.3 for those three datasets. There are three types of failure for the discovered equation.

| PDE | Observations | Method | Error ($E$) (low noise) | Error ($E$) (high noise) |
|---|---|---|---|---|
| Burgers' equation | 1000 (3.9%) | R-DISCOVER | **0.86±0.93%** | **2.28±0.82%** |
| | | DISCOVER | 1.85±6.07 | Fail (1)[a] |
| | | DeepMod | 2.55±0.13% | Fail (1) |
| | | PINN-SR | 0.88±0.03% | Fail (2)[b] |
| Nonlinear Fisher-KPP equation | 5000 (25%) | R-DISCOVER | **1.13±0.084%** | **14.12±14.31%** |
| | | DISCOVER | 25.20±0.29% | Fail (3)[c] |
| | | DeepMod | Fail (3) | Fail (3) |
| | | PINN-SR | Fail (3) | Fail (3) |
| KS equation | 20583 (20%) | R-DISCOVER | **1.15±1.86%** | **9.46±10.21%** |
| | | DISCOVER | Fail (3) | Fail (3) |
| | | DeepMod | 1.83±1.17% | Fail (3) |
| | | PINN-SR | Fail (1) | Fail (3) |

[a]Fail (1) occurs when all terms of the true equation are discovered, but redundant terms are also included. For Burgers' equation, for example $u_t = 0.066u_{xx} - 0.521uu_x + 0.1376uu_{xx} + 0.2637u^2$.

[b]Fail (2) denotes the failure that part of the true equation terms is identified without redundant terms. For Burgers' equation, for example $u_t = -0.394u_x$.

[c]Fail (3) represents a worse result, where not only are the correct terms missing, but incorrect terms are also included.

For Burgers' equation, for example $u_t = -0.667u_{xx}$.

Table VIII demonstrates that our framework can successfully uncover the correct PDE in both high-level noise and low-level noise cases. DISCOVER utilizes DNNs and sparsity penalties to ensure the derivation of concise and accurate equation terms, but is insufficient when dealing with higher levels of noise and less data amount. With the incorporation of discovered equations and a new model selection metric, the proposed framework is more robust and capable of addressing equation discovery tasks within systems characterized by greater complexity and stronger nonlinearity. The PINN-SR, which is sensitive to the hyperparameters of the sparsity penalty, tends to lack generalizability for equations without prior knowledge or with highly noisy data. DeepMod is also limited by the threshold setting and struggles to identify correct function terms for equations with high variance in coefficients, such as the nonlinear Fisher-KPP equation. Note that both methods exhibit poor performance in this equation, and another potential reason for this lies in the construction of the overcomplete function term library. To enable the discovery of $(u_x)^2$, we introduce additional function terms $(\partial_x^m u)^2$ and their polynomial combinations with other terms into the library. It significantly increases the difficulty of training and escalates the training cost. The specific equations discovered by the three methods are provided in the Supplementary material. FIG. 11 further offers a comprehensive comparison of the performance of R-DISCOVER with DeepMod under different noise levels, including metrics such as $L_2$, TPR, and $E_2$. PINN-SR is not considered here for further comparison since it tends to uncover redundant terms and its performance is unstable in the repetitive experiments, especially when higher noise is added. The comparison underscores that, even in the absence of prior knowledge, our method not only exhibits superior robustness to noisy data, but also boasts greater stability and general applicability.

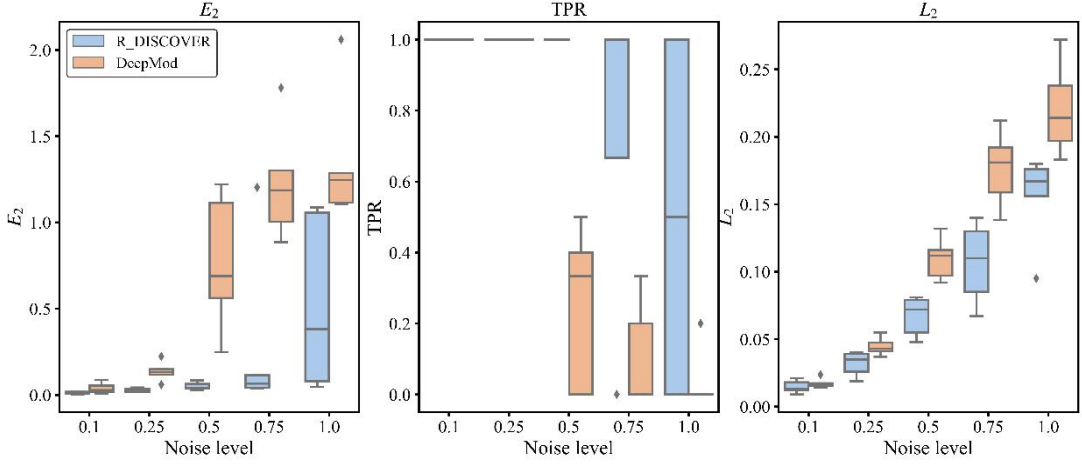

**FIG. 11.** Performance comparisons between our method with DeepMod on Burgers' equation. 1000 measurements are utilized and each experiment is repeated by 5 times with different random seeds.

## D. Effect of stability selection

At the end of the discovering process at each iteration, we select the top $K$ (default value is 3) expressions with the highest reward as the candidates for further evaluation. A higher reward indicates a smaller MSE between the right-hand side and left-hand side of the discovered equation. However, this metric is not adequate for determining the optimal equation representation due to potential overfitting issues in the presence of noise. Therefore, we need to conduct further model selection to identify the true PDE expression based on the performance of the stability evaluation. The Fisher-KPP equations are taken as examples for further illustration. Only the 1$^{st}$ round of discovery process is considered. As shown in FIG. 12, we provide the performance of the Top 3 expressions of classic Fisher-KPP under 100% noise added. It can be seen that the residuals (represented by MSE) of the first expression are smaller in different subsets, i.e., they conform to the data better. Nevertheless, the coefficients of the function terms are more drastically varied (the value of CV is large), and therefore it ranks lower in the final ranking of MSE×CV, which reveals that there may be an overfitting to the data. The specific statistics for the three expressions are shown in Table IX.

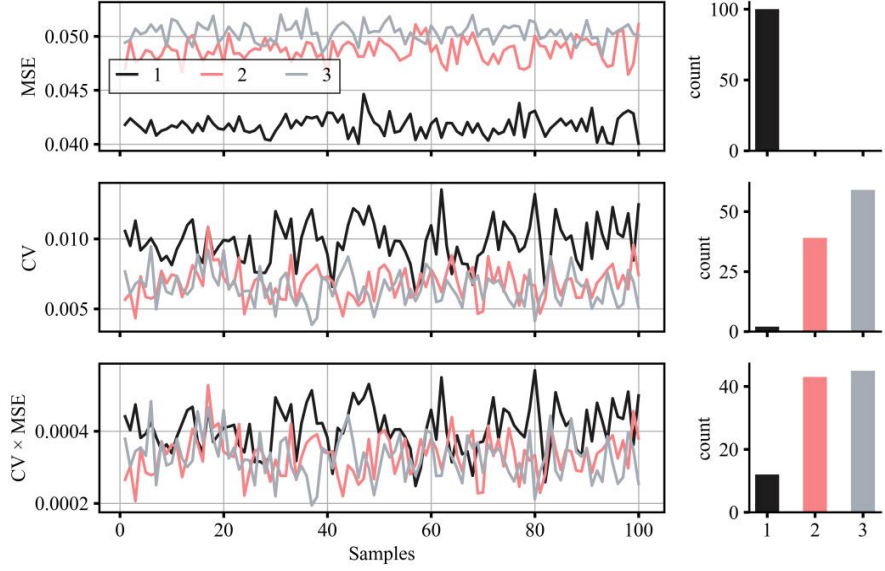

**FIG. 12.** Stability evaluation for classical Fisher-KPP equation

**TABLE IX.** Top 3 candidates for 1st round of searching classical Fisher-KPP equation.

| Candidates (Top 3) | Identified PDE | Ranking MSE×CV |
|---|---|---|
| 1 | $u_t = 0.023u_{xx} + 0.079u_x + 9.202u - 8.938u^2$ | 12 |
| 2 | $u_t = 0.025uu_{xx} + 0.025(u_x)^2 + 7.548u - 7.147u^2$ | 43 |
| 3 | $u_t = 0.023u_{xx} + 9.202u - 8.938u^2$ | **45** |

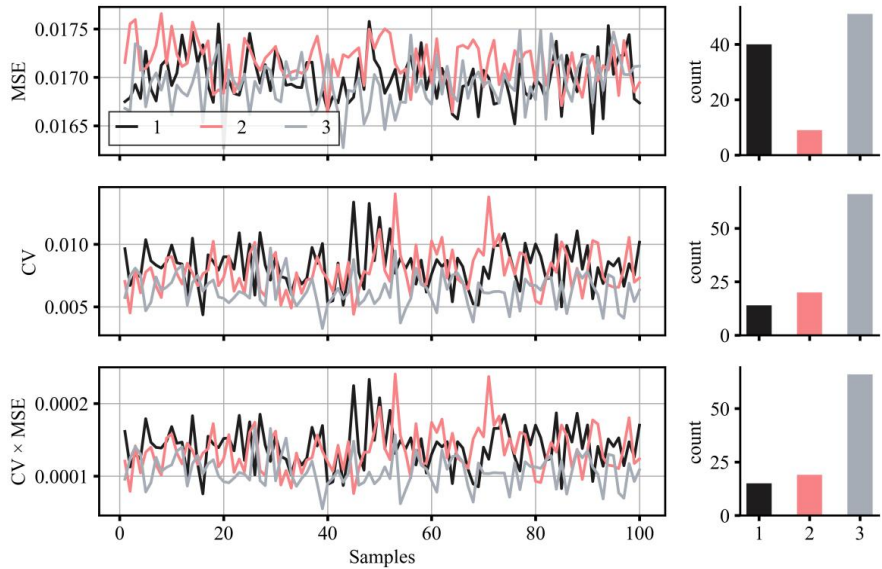

**FIG. 13.** Stability evaluation for Nonlinear Fisher-KPP equation.

**TABLE X.** Top 3 candidates for 1st round of searching nonlinear Fisher-KPP equation.

| Candidates (Top 3) | Identified PDE | Ranking MSE×CV |
|---|---|---|
| 1 | $u_t = 0.011uu_{xx} + 0.011(u_x)^2 + 9.844u - 9.896u^2 - 0.001(u-x)^2$ | 15 |
| 2 | $u_t = -0.06(u-x)^2 + 10.056u - 10.07u^2$ | 19 |
| 3 | $u_t = 0.012uu_{xx} + 0.012(u_x)^2 + 9.84u - 9.90u^2$ | **66** |

For the nonlinear Fisher-KPP equation with 50% noise added, expression 3 performs best in all three metrics, as shown in FIG. 13 and Table X. It can be seen that the performance of discovering equations on full data is insufficient and can only partially reflect the fitness of data. Notably, to ensure the robustness of the results, our final optimal equation is selected based on voting across 100 sample subsets, which to some extent mitigates the impact of uneven data distribution and noise. Experiments show that further stability evaluation is necessary and their combinations can comprehensively evaluate the discovered PDE candidates and have consistent effectiveness under different datasets, noise levels, and data volumes.

## IV. CONCLUSIONS

In this paper, we propose a framework named R-DISCOVER, which comprises two complementary and alternating update processes, including the discovering process and the embedding process. The open-form PDEs are generated with a RL-guided hybrid PDE generator and optimized with a RL strategy in the discovering process. Then the PDE with the best performance is embedded as a physical constraint into DNN during the embedding process. We validated this framework on several numerical experiments, which demonstrates that it is capable of identifying the correct governing equations of nonlinear dynamic systems from limited and highly noisy data. The novel hybrid PDE generator combines RL (neural guided), GA, and a dynamic subtree bank, which enables the framework to generate diverse PDEs and handle the discovery of lengthy and complex equations, such as the Navier-Stokes equations, and Gray–Scott reaction-diffusion model. Notably, the robustness of R-DISCOVER is further strengthened with a new model selection metric incorporated by balancing fitting accuracy and stability of coefficients. Compared with existing methods using an overcomplete candidate library (e.g., PINN-SR and DeepMod), R-DISCOVER offers superior flexibility since the dependence on prior knowledge is avoided with the symbolic representation. Meanwhile, it also demonstrates better universality and applicability compared to common sparsity-promoting methods relying on finetuning of hyperparameters. In the process of discovering governing equations, the proposed framework accomplishes the data interpretation and the construction of the prediction model simultaneously. The experimental results demonstrate the remarkable robustness of the proposed framework, which accurately identifies the correct equation terms while maintaining small coefficient errors across different noise levels, data volumes, and datasets. This enables our framework to better handle applications in complex real-world scenarios, where the quality of available observations may be inferior, and the physical processes described tend to exhibit stronger nonlinearity.

R-DISCOVER shows great potential for exploring real-world systems with limited understanding. However, some improvements need to be further made to better serve practical nonlinear systems. Firstly, dealing with complex multi-dimensional systems results in high computational costs for the predictive model due to the vast amount of data. The expressive ability of DNN proves to be limited, which necessitates the exploration of more powerful neural network structures. For example, convolutional neural networks could be employed to capture spatiotemporal features, which requires further comprehensive investigation, especially on the embedding method of physical information. Secondly, in the process of mining sparse and noisy governing equations, existing methods that rely on numerical differentiation and sparsity promotion tend to be more computationally efficient when applied to fixed candidate libraries, such as a series of studies of SINDy [1,12,35] and an integrated tool named PySINDy [56]. The proposed framework adopts symbolic mathematics and a RL-guided generator to increase the flexibility of PDE representations, enabling it to avoid the limitations of fixed libraries and uncover more complex structures. However, it simultaneously sacrifices efficiency to a certain degree. On one hand, the arbitrary form of generated PDEs enlarges the search space. On the other hand, numerically equivalent function terms may contain multiple different forms, such as $uu_x$ and $0.5(u^2)_x$, thereby increasing the difficulty of optimization. In the future, it is necessary to further integrate prior knowledge and necessary physical constraints to minimize the search space and time consumption during the discovery phase[57]. In the end, the current framework focuses on the identification of PDEs with constant coefficients in fluid dynamics. Symbolic representation has greater potential in discovering more complex equations in form, for example, parametric PDEs with time-dependant or spatial-dependant parameters[58,59] or fractional Schrödinger equation[60]. More importantly, the framework can be further improved and extended to cases in various scientific studies[61–63] and provide interpretable physical insights.

## ACKNOWLEDGMENTS

This work was supported and partially funded by the National Center for Applied Mathematics Shenzhen (NCAMS), the Shenzhen Key Laboratory of Natural Gas Hydrates (Grant No. ZDSYS20200421111201738), the National Natural Science Foundation of China (Grant No. 62106116) and the SUSTech – Qingdao New Energy Technology Research Institute.

## AUTHOR DECLARATIONS
### Conflict of Interest
The authors have no conflicts of interest to disclose.

## CODE AND DATA AVAILABILITY

The code of the framework and experimental data are available on GitHub at https://github.com/menggedu/DISCOVER.

# APPENDIX A: HYPERPARAMETERS

We conducted all the experiments with an NVIDIA A100 GPU card. The detailed hyperparameter setting in R-DISCOVER is provided in Table XI. It is worth noting that parameter tuning is comparatively straightforward. For governing equations of varying complexity, the hyperparameters of the hybrid PDE generators and NN-based reward evaluator need simple adjustments. For nonlinear systems with high dimensions or multiple state variables, the former necessitates an increase in the magnitude of $N$, i.e., generating more PDE samples in each iteration to produce diverse equation combinations and avoid falling into local optima. The latter requires an augmentation of the network complexity to improve its capacity to accurately fit the system response.

**TABLE XI**. Hyperparameters setting.

| Hyperparameter | Default value | Definition |
| --- | --- | --- |
| $N_s$ | 100 | Number of bootstrapped subsets |
| $N_p$ | 10 | Number of subsamples for calculating the coefficients of variation |
| $N_{sub}$ | 10 | The size of the basis function library subsampled from DSB |
| $N$ | 1000 | Number of the initial PDE traversals generated by the RL agent |
| $\lambda_1$ | 0.1 | Coefficients of physics loss term on collocation points |
| $\lambda_2$ | 0 | Coefficients of physics loss term on refined collocation points |
| $d_{max}$ | 4 | Maximum depth of subtrees |
| $\zeta_1$ | 0.01 | Penalty coefficients of the number of function terms |
| $\zeta_2$ | 0.0001 | Penalty coefficients of the maximum depth of subtrees |
| $\varepsilon$ | 0.02 | Threshold of reserved expressions |

# APPENDIX B: EXPERIMENT RESULTS

## 1. Reconstructed field

Our method can not only find the underlying governing equation from the noisy data, but also construct a prediction model for the nonlinear system response. With the discovered equation incorporated, it can provide more accurate predictions. The reconstructed fields predicted by the DNN for different systems are shown in FIG. 14, FIG. 15, and FIG. 16.

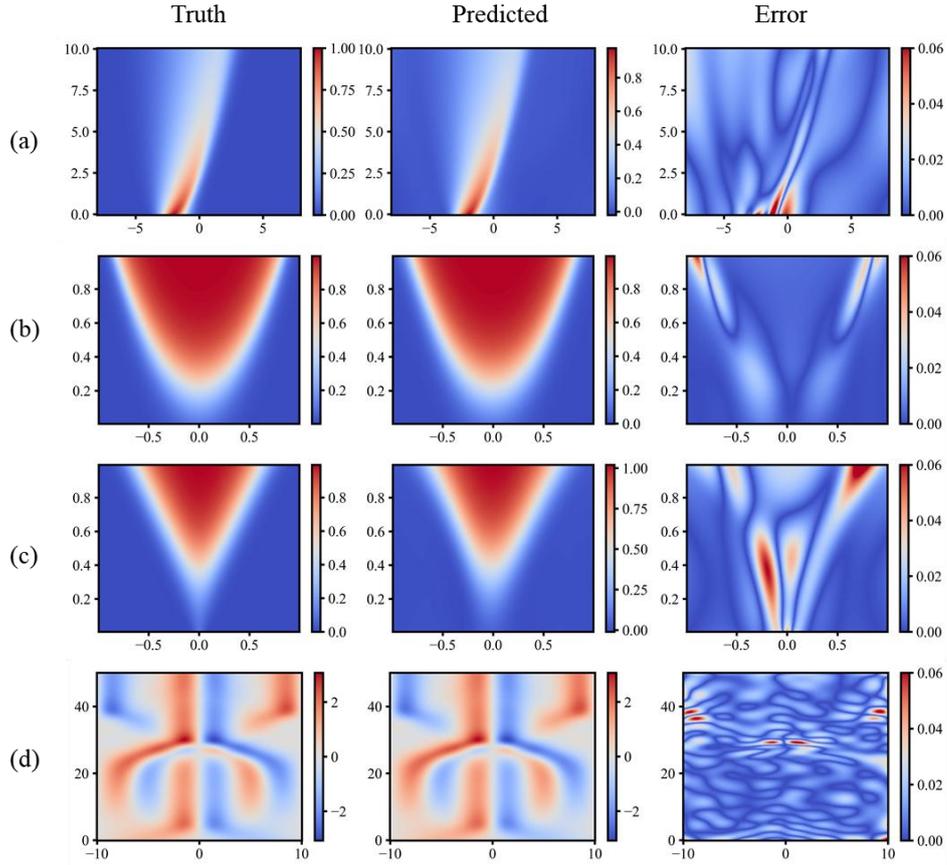

**FIG. 14.** Predicted solutions for (a) Burgers' equation with 100% noise and 1000 data points; (b) classical Fisher-KPP with 100% noise and 5000 data points; and (c) nonlinear Fisher-KPP equation with 50% noise and 10000 data points; (d) KS equation with 15% noise and 26214 data points.

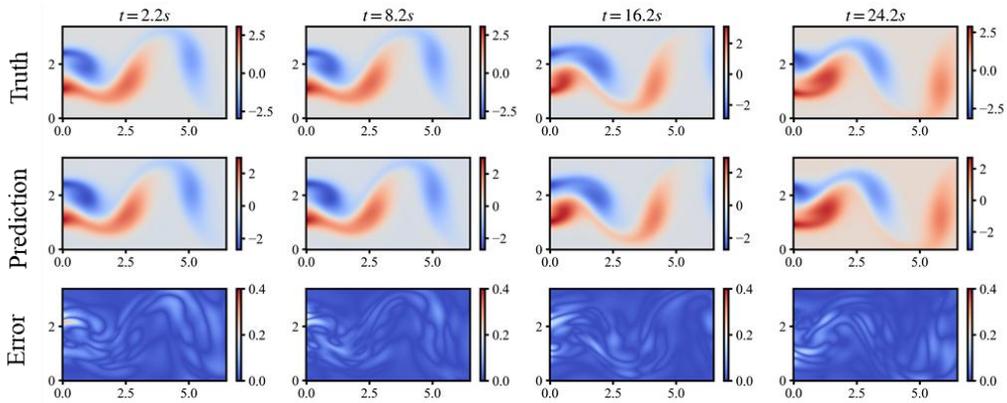

**FIG. 15.** Predicted solutions for the Navier-Stokes equation at different time steps with 25% noise and of total measurements.

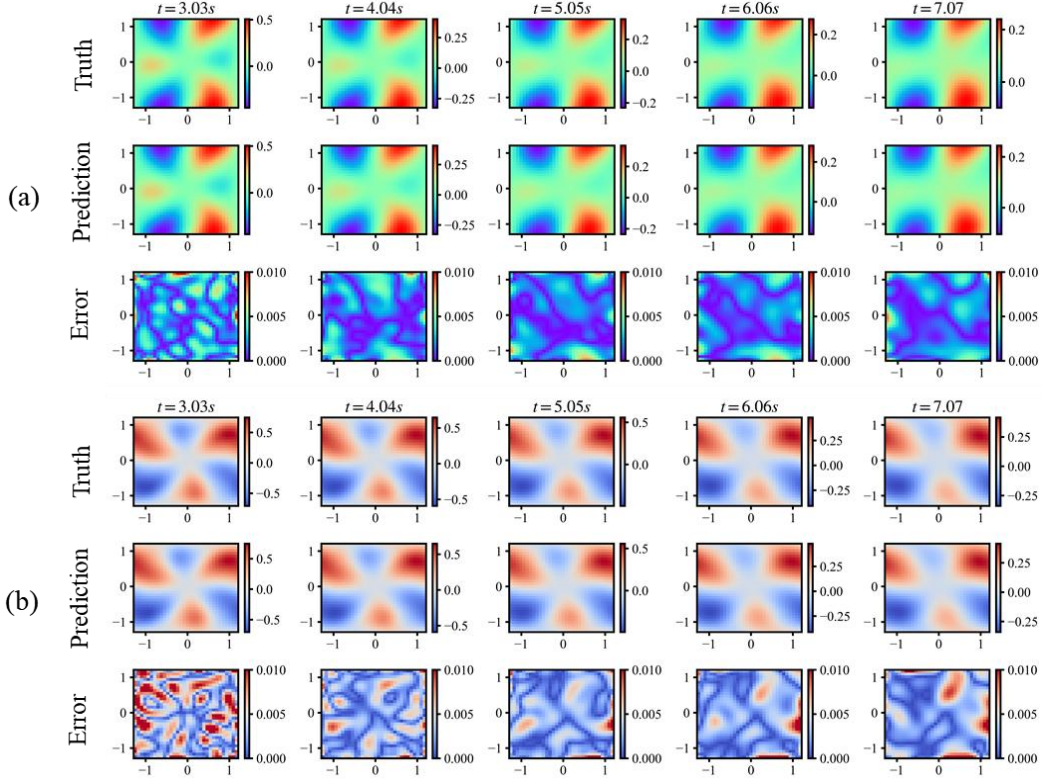

**FIG. 16.** Predicted solutions for a 3D reaction-diffusion model of (a) *u* and (b) *v* at different time steps with 20% noise and 20% of total measurements.

## 2. Comparison with PINN-based methods

We compared the performance of our model with our previous work-DISCOVER[10] and two PINN-based methods, including PINN-SR[24] and DeepMod[23] at three nonlinear governing equations. The concrete discovered equations under different levels of noise are shown in Table XII.

**TABLE XII.** The discovered equation by our framework and other related methods.

| PDE | Noise level | Method | Discovered equation |
|---|---|---|---|
| Burgers' equation | 10% | R-DISCOVER | $u_t = 0.0985 u_{xx} - 0.998 u u_x$ |
| | | DISCOVER | $u_t = 0.097 u_{xx} - 0.993 u u_x$ |
| | | DeepMod | $u_t = 0.095 u_{xx} - 1.004 u u_x$ |
| | | PINN-SR | $u_t = 0.099 u_{xx} - 0.986 u u_x$ |
| | 50% | R-DISCOVER | $u_t = 0.102 u_{xx} - 1.029 u u_x$ |
| | | DISCOVER | $u_t = 0.094 u_{xx} - 1.053 u u_x - 0.008 u^2_{xxx}$ |
| | | DeepMod | $u_t = 0.066 u_{xx} - 0.521 u u_x + 0.1376 u u_{xx} + 0.2637 u^2$ |
| | | PINN-SR | $u_t = -0.394 u_x$ |
| Nonlinear Fisher-KPP | 10% | R-DISCOVER | $u_t = 0.0196 u u_{xx} + 0.0196 u_x^2 + 10.029 u - 10.024 u^2$ |
| | | DISCOVER | $u_t = 0.010 u u_{xx} + 0.010 u_x^2 + 9.967 u - 9.954 u^2$ |
| equation | | DeepMod | $u_t = 0.634u_x^2 - 0.118uu_{xxx}$ <br> $+0.224u^2u_{xxx} - 0.124u^3u_{xxx}$ |
| | | PINN-SR | $u_t = 5.657u - 5.787u^3$ |
| | 30% | R-DISCOVER | $u_t = 0.015uu_{xx} + 0.015u_x^2 + 9.894u - 9.859u^2$ |
| | | DISCOVER | $u_t = 0.032u_x^2 + 9.066u - 9.007u^2 + 0.001$ |
| | | DeepMod | $u_t = 0.588u^2 + 2.967u - 0.106uu_{xxx}$ <br> $+0.208u^2u_{xxx} - 0.120u^3u_{xxx}$ |
| | | PINN-SR | $u_t = 5.821u - 5.901u^3$ |
| KS equation | 10% | R-DISCOVER | $u_t = -0.982uu_x - 0.984u_{xx} - 0.988u_{xxxx}$ |
| | | DISCOVER | $u_t = -0.0001u^4$ |
| | | DeepMod | $u_t = -1.015uu_x - 0.980u_{xx} - 0.992u_{xxxx}$ |
| | | PINN-SR | $u_t = -1.015uu_x - 1.00u_{xx} - 0.993u_{xxxx}$ <br> +7 redundant terms |
| | 30% | R-DISCOVER | $u_t = -0.941uu_x - 0.885u_{xx} - 0.879u_{xxxx}$ |
| | | DISCOVER | $u_t = -0.1628*uu_x - 0.1628u_xu_{xx}$ |
| | | DeepMod | $u_t = -1.189uu_x - 1.0951u_{xx} - 1.0760u_{xxxx}$ <br> $+0.236u^2u_{xx} + 0.183u^2u_{xx}$ |
| | | PINN-SR | $u_t = -0.865uu_x - 0.67u_{xxxx}$ <br> +5 redundant terms |